\documentclass[preprint,12pt]{elsarticle}

\usepackage{amssymb}
\usepackage{comment}
\usepackage{booktabs}
\usepackage{array}

\setcitestyle{numbers,square,comma}

\usepackage{amsmath}

\DeclareMathOperator*{\argmin}{arg\,min}
\usepackage{caption}

\usepackage{graphicx}
\usepackage{caption}
\usepackage{subcaption}
\usepackage{kantlipsum} %dummy text 
\usepackage{float}

\usepackage{amsmath} % for \mathbb
\usepackage{xcolor} % for \color

\definecolor{orange}{RGB}{255,165,0}

\usepackage{tabularray}
\usepackage[linesnumbered,ruled,vlined]{algorithm2e}
\usepackage{amsmath}
\makeatletter
\let\reftagform@=\tagform@
\def\tagform@#1{\maketag@@@{(\ignorespaces\textcolor{blue}{#1}\unskip\@@italiccorr)}}
\renewcommand{\eqref}[1]{\textup{\reftagform@{\ref{#1}}}}
\makeatother
\usepackage{hyperref}
\hypersetup{colorlinks,allcolors=black}

\usepackage{lineno}
\usepackage{subcaption}
\newcommand{\minus}{\scalebox{0.75}[1.0]{$-$}}

\usepackage{geometry}

\usepackage{graphicx}
\usepackage{caption}
\usepackage[skip=0.5ex]{subcaption}
\usepackage[parfill]{parskip}

\begin{document}

\begin{frontmatter}



\title{Scaling up ridge regression for brain encoding in a massive individual fMRI dataset}
\title{Scaling up ridge regression for brain encoding in a massive individual fMRI dataset}
\author[label1,label2,label3]{Sana Ahmadi\footnote{Corresponding author,   sana.ahmadi@mail.concordia.ca}}
\author[label2,label3]{Pierre Bellec\footnote{pierre.bellec@criugm.qc.ca}}
\author[label1]{Tristan Glatard\footnote{tristan.glatard@concordia.ca}}
\address[label1]{Department of Computer Science and Software Engineering, Concordia University, Montreal, QC, Canada}
\address[label2]{Université de Montréal, Montréal, QC, Canada}
\address[label3]{Centre de Recherche de L'Institut Universitaire de Gériatrie de Montréal, Montréal, Canada}


%\address{California, United States}

\begin{abstract}
Brain encoding with neuroimaging data is an established analysis aimed at predicting human brain activity directly from complex stimuli features such as movie frames. Typically, these features are the latent space representation from an artificial neural network, and the stimuli are image, audio or text inputs. Ridge regression is a popular prediction model for brain encoding due to its good out-of-sample generalization performance. However, training a ridge regression model can be highly time-consuming when dealing with large-scale deep functional magnetic resonance imaging (fMRI) datasets that include many space-time samples of brain activity. This paper evaluates different parallelization techniques to reduce the training time of brain encoding with ridge regression on the CNeuroMod Friends dataset, one of the largest deep fMRI resource currently available. With multi-threading, our results show that the Intel Math Kernel Library (MKL) significantly outperforms the OpenBLAS library, being 1.9 times faster using 32 threads on a single machine. Yet, the performance benefits of multi-threading are limited, and reached a plateau after 8 threads in our main experiment. We then evaluated the Dask multi-CPU implementation of ridge regression readily available in scikit-learn (MultiOutput), and we proposed a new ``batch'' version of Dask parallelization, motivated by a time complexity analysis. With this Batch-MultiOutput approach, batches of brain targets are processed in parallel across multiple machines, and multi-threading is applied concurrently to further accelerate computation within a batch. In line with our theoretical analysis, MultiOutput parallelization was found to be impractical, i.e., slower than multi-threading on a single machine. In contrast, the Batch-MultiOutput regression scaled well across compute nodes and threads, providing speed-ups of up to 33$\times$ with 8 compute nodes and 32 threads compared to a single-threaded scikit-learn execution. Batch parallelization using Dask thus emerges as a scalable approach for brain encoding with ridge regression on high-performance computing systems using scikit-learn and large fMRI datasets. These conclusions likely apply as well to many other applications featuring ridge regression with a large number of targets. 
\end{abstract}

\end{frontmatter}


\section{Introduction}
\label{S:1}
The human brain is a computing system with billions of neurons as computing units. Cognitive neuroscience aims to discover functional principles of brain organization by leveraging large-scale neuroimaging data. One of the key methods used for this purpose is brain encoding \cite{Encoding1}, in which a model predicts brain responses directly from rich stimuli such as natural images or videos, using the internal representations of an artificial neural network as a feature space for prediction.

Among the regression methods used in brain encoding to predict brain activity, ridge regression~\cite{ridge1} has become popular and well-accepted~\cite{Leila_work, NLP_models, newTvision, GPT_breain, Lanq-vision, bertmodel,ref_ridge1,ref_ridge2,ref_ridge3, VR3} due to its two key features: 1) ridge regression tends to be generalizable to new stimuli and avoids overfitting, and 2) efficient implementations of ridge regression are available ~\cite{banded1} which are less computationally intensive than other approaches. 

For brain encoding of visual tasks, ridge regression is often applied to the activations produced by various neural networks architecturesin response to visual stimuli, such as convolutional neural networks (CNN) and transformers ~\cite{kati, vision2, vision3, H_DS, BTB, LSTM1, IM_1}. For instance, in~\cite{VR3}, the authors compared the activation of CNN units with brain response to a dynamic visual stimulus (movie frames) and found that these representations were able to accurately predict fMRI data collected with human subjects watching movies.

Even though linear algebra optimizations exist for ridge regression~\cite{banded1}, the training process still requires compute-intensive matrix computations over the whole dataset. This computational cost is especially substantial for brain encoding models trained separately for each spatial measurement sample (voxel), as the number of voxels can range from tens to hundreds of thousands in a full brain fMRI acquisition with high spatial resolution. Thus, full brain encoding using ridge regression remains a challenge, even with modern computational resources. 

Furthermore, the computational requirements of ridge regression are exacerbated by the need to train brain encoding tasks on large datasets. Indeed, finding the correlation between the huge feature space of natural images and brain activity requires to explore a large space of visual stimuli~\cite{Dataset_p}. Over the past few years, the quantity and quality of fMRI datasets have increased rapidly in terms of the number of human subjects, the number of scanning hours available for each subject, as well as spatio-temporal resolution. In particular, datasets such as BOLD5000 \cite{Bold5000}, Natural Scenes Dataset (NSD)\cite{NSD_dataset} provide so-called deep fMRI datasets, with long scanning time for a few subjects and an extensive stimuli space to properly estimate the generalization of brain encoding to different types of stimuli, e.g. images from many different categories. Training purely individual brain models also by-pass the challenges of modelling inter-individual variations in brain organization, which is substantial \cite{Gratton}. As a consequence of increased spatial resolution and volume of time samples available for a single subject with the advance of simultaneous multislice fMRI REF, there is thus an urgent need to understand the efficiency of various implementations of ridge regression for brain encoding with large fMRI datasets.

The CNeuroMod research group \cite{cneuromod}  has released the largest fMRI dataset for individual brain modeling currently available, featuring up to 200 hours of fMRI data per subject (N=6).  The CNeuromod dataset provides an opportunity to train complex brain encoding models based on artificial neural networks, but also raises substantially the computational costs of brain encoding. This work investigates several parallelization techniques for ridge regression, using the CNeuroMod Friends dataset to predict brain activity from video stimuli. We focus on a standard brain encoding pipeline using an established pretrained network (VGG16), and we used the scikit-learn library~\cite{pedregosa2011scikit} for brain encoding, that provides efficient implementations of various machine-learning models, including ridge regression. 
 
We benchmarked the efficiency of different types of parallelization, namely multi-threading (multiple cores on a single CPU) and multi-processing (distributing computations across multiple CPUs in a high performance computing environment). For multi-threading, scikit-learn can leverage the BLAS (Basic Linear Algebra Subprograms) specification for linear algebra implemented using the open-source OpenBLAS  library~\cite{OpenBLAS} or the proprietary Intel oneAPI Math Kernel Library (MKL)~\cite{MKL}. Both of these linear algebra libraries support multi-threading on a single CPU. Moreover, scikit-learn models rely on the \href{https://joblib.readthedocs.io}{Joblib library} to interface with various parallelization backends including Dask~\cite{Dask}, which can be used to distribute computations across multiple compute nodes. Specifically, we utilized scikit-learn's MultiOutput regressor, which by default trains individual ridge regression models for each target variable (here, each location in the brain) independently. The MultiOutput however comes with substantial overhead, as it introduces many redundant computations across brain targets. To reduce the amount of redundant computations happening with MultiOutput ridge regression, we also modified MultiOutput to train a series of model on batches of brain targets, using one compute node per batch and multi-threading execution within each batch. We conducted a theoretical complexity analysis to motivate the choice of this approach. We also repeated our benchmark for both MultiOutput and the batch MultiOutput by assessing the efficiency of parallelization with varying number of threads per node and the number of compute nodes. 

Taken together, this study will provide concrete guidelines for practitioners who want to run brain encoding efficiently with ridge regression and large fMRI datasets, using high-performance computing infrastructure and CPUs.

\section{Materials and Methods}
 
\subsection{fMRI dataset}

We used the 2020-alpha2 release of the Friends fMRI dataset collected by the Courtois project on neuronal modeling, CNeuroMod \cite{cneuromod1}. Some of the text in this section is adapted from the Courtois NeuroMod technical documentation (\url{https://docs.cneuromod.ca}).
 
\subsubsection{Friends TV show stimuli}
Participants watched three seasons of the Friends TV show while their brain activity was recorded using fMRI. Each episode was divided into two segments (a/b) to provide shorter scanning runs and allow participants to take a break. There was a slight overlap between the end of each video segment and the beginning of the next video segment to provide an opportunity for participants to catch up with the story line.
 
\subsubsection{Participants}
The Friends dataset includes fMRI time series collected on six participants in good general health, 3 women (sub-03, sub-04, and sub-06) and 3 men (sub-01, sub-02, and sub-05). Three of the participants reported being native francophone speakers (sub-01, sub-02, and sub-04), one as being a native anglophone (sub-06), and two as bilingual native speakers (sub-03 and sub-05). All subjects had a good comprehension of English, which was used in the sound track of the Friends videos. All subjects also provided written informed consent to participate in this study, which was approved by the local research ethics review board (under project number CER VN 18-19-22) of the CIUSSS du Centre-Sud-de-l'Île-de-Montréal, Montréal, Canada.
 
\subsubsection{Magnetic resonance imaging}
Magnetic resonance imaging (MRI) was collected using a 3T Siemens Prisma Fit scanner and a 64-channel head/neck coil, located at the Unit for Functional Neuroimaging (UNF) of the Research Centre of the Montreal Geriatric Institute (CRIUGM), Montréal, Canada. Functional MRI data were collected using an accelerated simultaneous multi-slice, gradient echo-planar imaging sequence \cite{Setsompop,Xu} developed at the University of Minnesota, as part of the Human Connectome (HCP) Project \cite{Glasser}. The fMRI sequence used the following parameters: slice acceleration factor = 4, TR = 1.49s, TE = 37 ms, flip angle = 52 degrees, 2 mm isotropic spatial resolution, 60 slices, acquisition matrix 96x96. The structural data was acquired using a T1-weighted MPRAGE 3D sagittal and the following parameters: duration 6:38 min, TR = 2.4 s, TE = 2.2 ms, flip angle = 8 deg, voxel size = 0.8 mm isotropic, R=2 acceleration.  
For more information on the sequences used or information on data acquisition (including fMRI setup), visit the \href{https://docs.cneuromod.ca/en/latest/MRI.html#sequences} {CNeuroMod technical documentation} page. 

\subsubsection{Preprocessing}
All fMRI data were preprocessed using the fMRIprep pipeline version 20.2.3 \cite{Esteban}. We applied a volume-based spatial normalization to standard space (\allowbreak{MNI152 NLin2009cAsym}). Furthermore, a denoising strategy was applied to regress out the following basic confounds: (1) a 24-degrees of freedom expansion of the motion parameters, (2) a basis of slow time drifts (slower than 0.01 Hz). This step was implemented with the Nilearn maskers (see below) and the \texttt{load\_confounds} tool\footnote{\url{https://github.com/simexp/load_confounds}} (option \texttt{Params24}). A spatial smoothing with a 8 mm field-width-at-half-maximum and a Gaussian kernel was also applied with Nilearn prior to time series extraction. For each fMRI run, time series were also normalized to zero mean and unit variance (over time, for each voxel independently).

\subsubsection{Multiresolution time series extraction}
\begin{table}[ht]
\small
\begin{center}
\captionof{table}{Brain datasets summary: number ($n\times t$) of time x space samples and size (in MB or GB) of fMRI time series in three resolutions.\label{Tab:Tcr}}
\begin{tabular}{l l r r r} 
Resolution & Subject & $n$ & $t$  & Size (float64)\\ [0.5ex] 
\hline
Parcels      & sub-0(1-6)& 69,202 & 444 & 244 MB \\
ROI         & sub-0(1-6) & " & 6,728 & 2.6 GB \\
Whole-Brain & sub-01      & " & 264,805 & 138 GB\\
            & sub-02      & " & 266,126 & 142 GB\\
            & sub-03      & " & 261,880 & 136 GB\\
            & sub-04      & " & 266,391 & 142 GB\\
            & sub-05      & " & 263,574& 138 GB\\
            & sub-06      & " & 281,532 & 148 GB\\
Whole-Brain (B-MOR) & sub-01      & 10,000 & 264,805 &  21 GB\\
            & sub-02      & " & 266,126 &  21.2 GB\\
            & sub-03      & " & 261,880 & 20.8 GB\\
            & sub-04      & " & 266,391 & 21.2 GB\\
            & sub-05      & " & 263,574&  21 GB\\
            & sub-06      & " & 281,532 & 21.8 GB\\

Whole brain (MOR)  
            & sub-0(1-6)      & 1,000 & 2,000 &  16 MB\\

\hline
\end{tabular}
\end{center}
\end{table} % Table Tab:Tcr

Functional MRI data takes the form of a 3D+t array, where the 3D spatial dimensions encode for different spatial locations on a regular 3D sampling grid (with 2 mm isotropic voxels for this dataset) within the field of view of acquisition, and the time axis (t) encodes brain samples recorded at different times, again on a regular sampling grid (with the time interval TR=1.49s for this dataset). It is common practice to translate this 3D+t array into a 2D array, where the first dimension encodes time, and the second dimension encodes space. There are multiple ways to perform this translation, which corresponds to different spatial resolution choices for the analysis. In this work, we used the so-called maskers of the Nilearn library \cite{ML_neuroimaging} to perform this operation, and we considered three common spatial resolutions to investigate the scalability of different implementations of ridge regression. These approaches vary markedly in the size of the resulting spatial dimension: parcel-wise, ROI-wise, and whole brain. These three resolutions are further described below:
\begin{enumerate}
    \item \textbf{Parcels}: The preprocessed BOLD time series were averaged across all voxels in each parcel of a parcellation atlas, using the \texttt{NiftiLabelsMasker} masker from Nilearn. We used the Multiresolution Intrinsic Segmentation Template (MIST) \cite{MIST}.  MIST provides a hierarchical decomposition of functional brain networks in nine levels (7 to 444), and we used here the largest available resolution (444 brain parcels). 
    \item \textbf{ROI}: In this approach, a binary mask of the visual network was extracted from MIST at resolution 7. Voxel-wise time series were extracted for all voxels present in this mask, using the \texttt{NiftiMasker} masker from Nilearn. Note that the location of the mask was based on non-linear registration only, and did not use subject-specific segmentation of the grey matter. The exact same number of voxels (6728) was thus present in the mask for all subjects' data, after realignment in stereotaxic space. 
    \item \textbf{Whole-Brain}: In this approach, the brain mask generated by the fMRIprep pipeline based on the structural scans of each participant was resampled at the resolution of the fMRI data. This mask included both grey matter, white matter, cerebro-spinal fluid but excluded all tissues surrounding the brain. Voxel-wise time series of all voxels included in the mask were extracted, again using the \texttt{NiftiMasker} masker from Nilearn. As the brain mask was subject specific, the number of voxels in the mask varied slightly across subjects. 
   
\end{enumerate}
   
Table \ref{Tab:Tcr} presents the shape of the brain data array $Y$ with these three levels of resolution and six subjects, where the number of rows and columns indicate the number of volumes ($n$ time sample) and targets ($t$ spatial targets) respectively. The temporal dimension is identical for all three approaches, while the spatial dimension of ROI is one order of magnitude larger than Parcels, and the spatial dimension of Whole-Brain is three orders of magnitude larger than Parcels. 
We also introduce two truncated versions of the whole-brain resolution, marked as MOR and B-MOR in Table \ref{Tab:Tcr}, which represent subsets of the dataset. We truncated the number of time samples and brain target from the whole-brain data, to accommodate memory requirements in the benchmark infrastructure. In this table,  memory sizes are presented in the float64 format used in Scikit-learn for ridge regression.

\subsection{Brain encoding}
\label{vgg-explanation}
\begin{figure}[t]
%\hspace* {- 1 cm}
\centering
\includegraphics[width=14cm]{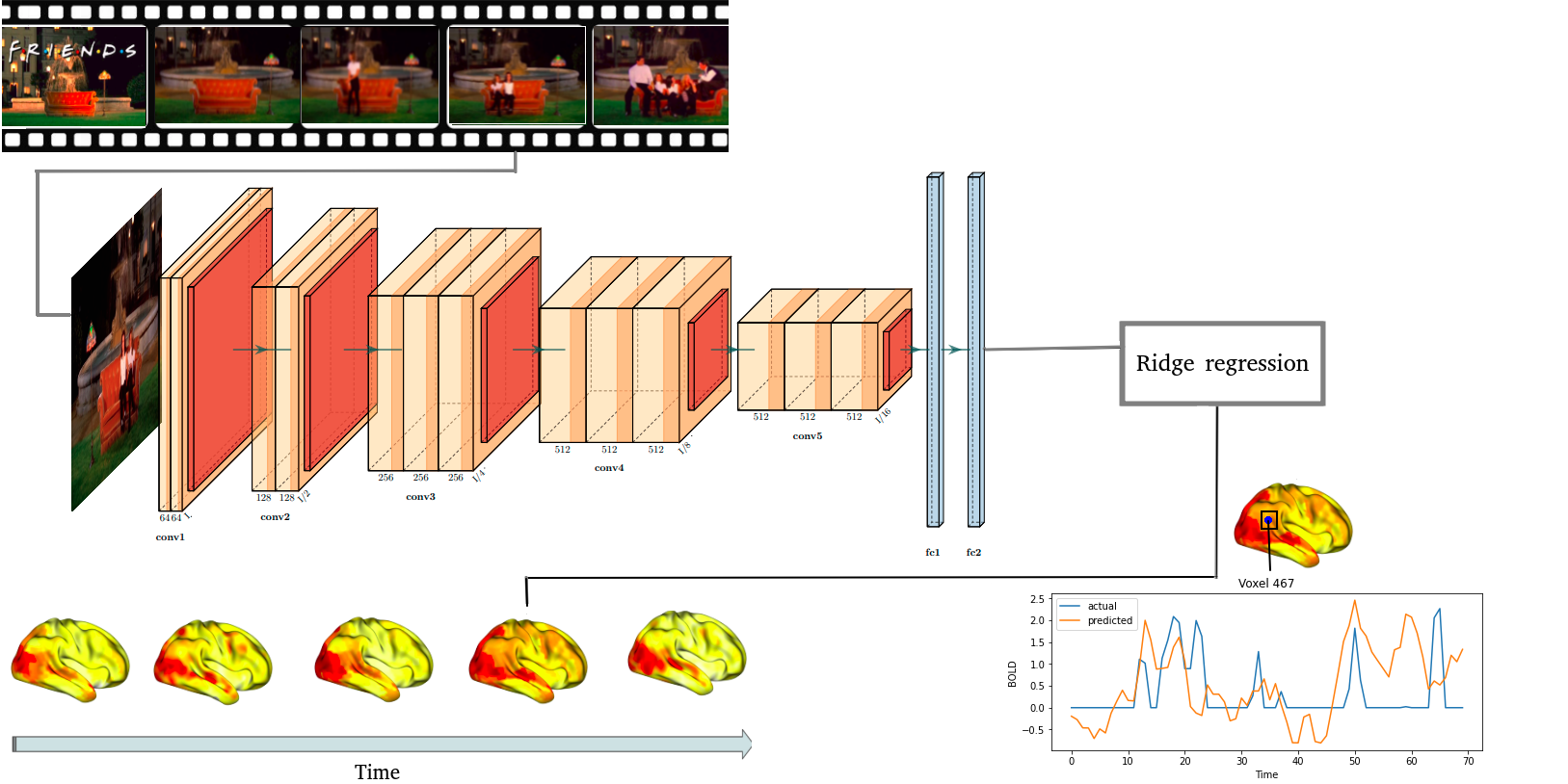}
\caption{The two main steps of brain encoding: Extracting features from movie frames using VGG16 pretrained model and predicting brain response using ridge regression.}
        \label{Brain encoding Fig}

\end{figure}

Figure~\ref{Brain encoding Fig} recapitulates the two main steps of brain encoding: extracting features from movie frames through a pretrained artificial network (here VGG16) and predicting brain response using a regularized linear regression model, called ridge. The ridge regression is trained through pairs of prediction targets (fMRI data $Y$) and dynamic visual stimuli features (predictors $X$), and experiments are implemented at several levels of resolution, see Table \ref{Tab:Tcr} to test the scaling efficiency of different implementations of the ridge regression.
 
\subsubsection{VGG16 artificial vision network}
In this work, we used the approach of \cite{VR3,conwell2022pressures}, and applied a VGG16 model \cite{VGG16_ref} pretrained for image classification to extract visual features from the movie frames. The VGG16 model was trained on a dataset of over 2 million images belonging to 1000 classes from the ImageNet database \cite{ImageNet}, and the weights of the models were retrieved through  \href{https://www.tensorflow.org/} {TensorFlow}. This model achieved 92.7\% top-5 test accuracy for image-object classification. The VGG16 architecture \cite{vgg16} is a widely-used convolutional neural network (CNN) known for its simplicity and effectiveness in image classification tasks \cite{conwell2022pressures}.  The network comprises 16 layers, including 13 convolutional layers and 3 fully connected layers. The convolutional layers use small 3x3 filters with a stride of 1 and employ rectified linear unit (ReLU) activation functions. Max-pooling layers with 2x2 filters are applied for spatial down-sampling. The architecture is characterized by a large number of trainable parameters, summarized in Appendix \ref{Appendix1} (based on the TensorFlow summary of the model), making it suitable for various computer vision applications. 

\subsubsection{Extracting VGG16 features of dynamic visual stimuli}  
For each of the $n=69,202$ fMRI time samples, we extracted the stimulus video frames corresponding to the 4 TRs immediately preceding each fMRI samples (equivalent to a window of 4 x 1.49 = 5.96s duration). This operation was done to take into account the known delayed, convolutional nature \cite{Neurophysiological} of the relationship between the visual stimulus and the hemodynamic response. Each frame was resampled to a (224, 224, 3) array and fed into VGG16 to extract 4096 features from the last layer. The features of the 4 TRs were concatenated, resulting into a single feature vector of length $p=16384$. In total, the array of features $X$ used for brain encoding had a size of $(n=69202, p=16384)$ (number of time samples x number of features), or 8.5 GB (in float32 precision). 

\subsubsection{Ridge regression}
\label{sec:ridge}
Ridge regression was first proposed by Hoerl and Kennard~\cite{ridge1} as a generalization of ordinary least-square regression. In comparison with ordinary least mean square regression, ridge regression provides better generalization to unseen data through regularization of coefficient estimates, in particular in the presence of a large number of predictor variables. The ridge regression is expressed as the following optimization problem solving for regression coefficients $b^{\ast}$ independently at each spatial location:
\begin{equation}
     b^{\ast} = \argmin_{b \in \mathbb{R} ^{p}} \left( \Vert y - Xb \Vert_{2} ^ {2} +  \lambda  \Vert b\Vert _{2} ^{2}\right), \label{eq:ridge}
\end{equation}
where $X \in \mathbb{R} ^{n \times p}$ is the matrix of stimuli features with $n$ time samples and $p$ features, $\Vert .\Vert _{2}$ is the $ \boldsymbol\ell^2$ norm of a vector, and $y \in \mathbb{R} ^ n$ is the target vector obtained from fMRI data at a single spatial location (at either Parcels, ROI or Whole-Brain resolutions). The hyper-parameter $\lambda$ is used to control the weighting of the penalty in the loss function. The best value for $\lambda$ is estimated among a set of candidate values through cross-validation, as explained below. If the value of $\lambda$ is too low the training process may overfit and if the value of $\lambda$ is too high then the brain encoder model may underfit~\cite{banded1}. 

\subsubsection{Brain encoding performance and hyper-parameter optimization}
For a given subject, the samples $X$ were split into training (90\% random) and test (10\% remaining) subsets. The coefficients of the ridge regression were selected through Eq. \ref{eq:ridge} based on the training set only. Table \ref{Tab:weight_matrix} presents the memory size and number of ridge training parameters with three levels
of resolution across six subjects. 

\begin{table}[ht]
\footnotesize
\begin{center}
\captionof{table}{Number of training parameters (rounded to closest M) and size of weight matrices in three resolutions for brain encoding \label{Tab:weight_matrix}}
\begin{tabular}{llrr} 
Resolution & Subject &  \# of training & Size(float64)\\ 
           &         & parameters  &\\
\hline
Parcel      & sub-0(1-6)& 7 M &  58 MB \\
ROI         & sub-0(1-6) & 110 M &  1.2 GB \\
Whole brain and Whole brain (B-MOR)  & sub-01      & 4338 M & 34.6 GB\\
  & sub-02      & 4360 M &  34.8 GB\\
            & sub-03      & 4290 M &  34.2 GB\\
            & sub-04      &  4364 M &  34.6 GB\\
            & sub-05      & 4318 M &  34.6 GB\\
            & sub-06      & 4612 M &  34.8 GB\\
 Whole brain (MOR)
            &sub-0(1-6)      & 32.7 M & 262.0 MB \\
\hline
\end{tabular}
\end{center}
\end{table} % Table Tab:Tcr

We measured the final quality of brain encoding as the Pearson's correlation coefficient between the actual fMRI time series and the time series predicted by the ridge regression model, on the test set. A leave-one-out validation was used inside the training set to estimate the hyper-parameter value $\lambda$ with optimal performance (based on cost function defined in Eq. \ref{eq:ridge}), based on the grid:
$$\lambda \in \left\{0.1, 1, 100, 200, 300, 400, 600, 800, 900, 1000, 1200 \right\}.$$
The choice of $\lambda$ can either be made separately for each of the $t$ brain targets, or a common value can be selected for all brain targets based on the average performance of the model across all $t$ brain targets. In this work, a single $\lambda$ is used for all targets.

\subsection{Ridge regression implementations} 
\subsubsection{Scikit-learn efficient ridge implementation}
\label{sec:efficient-implementation}
In large-scale brain encoding tasks, the computational cost of ridge regression increases linearly with the increasing number of targets. To reduce computing time when multiple targets are used, formulations of ridge regression have been proposed to mutualize computations among the targets. The formulation described below was presented in~\cite{banded1} and is implemented in the scikit-learn library.

In a multi-target regression problem, vector $y$ in Equation~\ref{eq:ridge} is now a matrix $Y$ of size $n$ by $t$ where $t$ is the number of spatial targets. Matrix $X$ is still of dimension $n$ by $p$.  The weight matrix $W$ can be calculated as follows: 
 \begin{equation}
 \label{eq:weight_computation}
W=MY
\end{equation}
where
 \begin{equation}
 \label{eq:M_computation}
M = (X^T X + \lambda I_{p}) ^{\minus 1}  X^T
\end{equation}
and $I_{p}$ is the identity matrix. The key point is that $M$ is independent of Y and can therefore be reused for all $t$ targets. This strategy reduces the time complexity of multi-target ridge regression from $O(p^{3}t + p^{2}nt) $ to  $O(p^3 + p^2 n + pnt)$~\cite{banded1}, see Section \ref{sec:complexity} for details.

 Scikit-learn also mutualizes computations among subsequent estimations of M for different $\lambda$ values, typically encountered during hyper-parameter optimization. To do so, it relies on the SVD decomposition of $X$:
 \begin{equation}
     X = USV^T,
 \end{equation}
 where $U \in \mathbb{R} ^{n \times p}$ and $V \in \mathbb{R} ^{p \times p}$ are orthonormal matrices, and $S \in \mathbb{R} ^{p \times p}$ is diagonal.  Then, the matrix $M$ can be rewritten as: 
  \begin{equation}
 \label{eq:M_svd}
 M (\lambda)= V (S^2 + \lambda I_{p}) ^ {\minus1} S U^{T}
 \end{equation}
 
 Computing $(S^2 + \lambda I_{p}) ^ {\minus1} S$ is inexpensive as this matrix is diagonal. SVD decomposition of feature matrix $X$  reduces complexity of computing M  from $O(p^{3}r + pnr)$ to $O(p^{2}nr)$~\cite{banded1}, where $r$ is the number of tested hyper-parameter values (see Section \ref{sec:complexity} for details).

\subsubsection{Computational environment}
Brain encoding experiments were run on Beluga, a high-performance computing (HPC) cluster of Canada Digital Alliance, providing researchers with a robust infrastructure for advanced scientific computations. Beluga features numerous compute nodes, high-speed interconnects, and parallel processing capabilities, visit the \href{https://docs.alliancecan.ca/mediawiki/index.php?title=B%C3%A9luga/en} {Beluga technical documentation} page for details. 

All benchmarking experiments were run on a high-performance computing cluster called ``slashbin", fully dedicated to benchmarking, without concurrent users accessing the platform during tests. This cluster was located at Concordia University Montreal, and featured 8 compute nodes. Each compute node featured an Intel\textregistered Xeon\textregistered Gold 6130 CPU @ 2.10GHz 
with 32 physical cores (64 hyperthreaded cores), 250~GB of RAM, Rocky Linux 8.9 with Linux 
kernel 4.18.0-477.10.1.el8\_lustre.x86\_64. Input and output data were located on a 960GB Serial-Attached SCSI (SAS) 12GBPS 512E Solid State Drive that was network mounted to each compute node using NFS v4.

\subsubsection{Multi-threading parallelism}

\begin{figure}[ht]
%\hspace*{-4.5cm}
     \centering
    
\includegraphics[width= 0.6\textwidth]{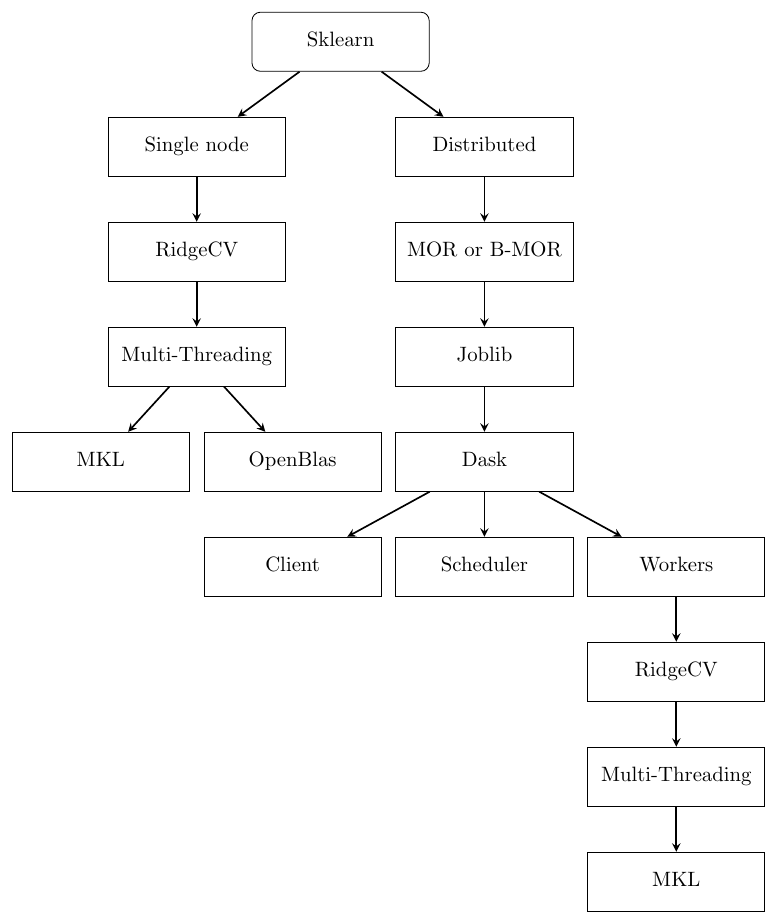}                
       \caption{Mutilthreading and Distributed parallelism in scikit-learn's ridge regression}
        \label{flochart}
\end{figure}

Multi-threading is a mechanism to parallelize executions on multi-core CPUs. In the case of ridge regression, multi-threading is available mainly for linear algebra routines implemented through the Basic Linear Algebra Subprograms (BLAS) specification. Two well-known BLAS libraries are Intel Math Kernel Library (MKL) \cite{MKL} and OpenBLAS  \cite{OpenBLAS}. These BLAS libraries incorporate optimized implementations that leverage multi-threading parallelism for efficient execution. In particular, the OpenBLAS  and MKL libraries enable multithreaded execution of ridge regression over the CPU cores in a single machine for a faster execution time. Figure~\ref{flochart} summarizes the different parallelization modes benchmarked by our experiments.

\subsubsection{Distributed parallelism}
\label{Section Distributed parallelism}
\begin{figure}[tp]
%\hspace*{-1.57cm}
\includegraphics[width=14cm]{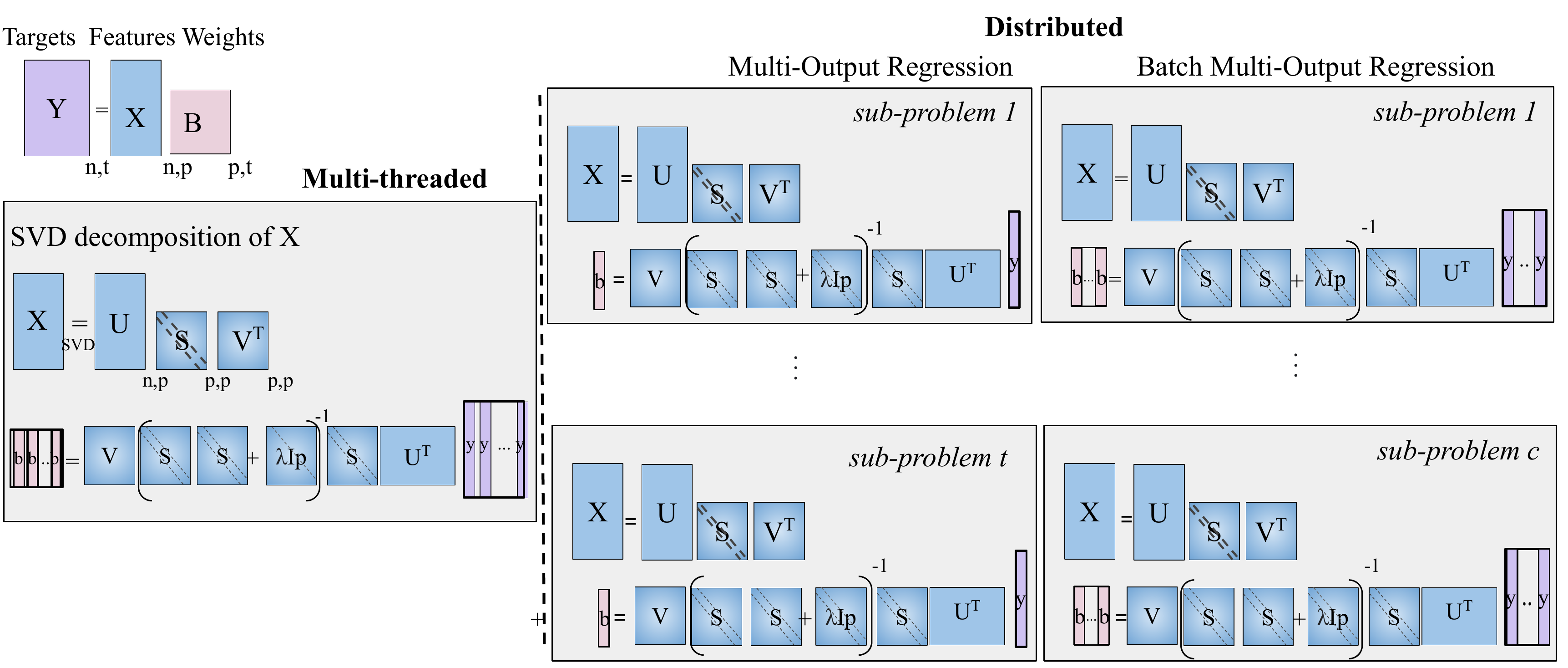}
\centering
\caption {Matrix computations in Multi-threading ridgeCV,  MOR and B-MOR model fitting. Assuming  $X \in \mathbb{R} ^{n \times p}$,  $Y \in \mathbb{R} ^ {n \times t}$  and $X= USV^T$ then the weight matrix  $B  \in \mathbb{R} ^ {p \times t}$ 
equals to $B = V (S^2 + \lambda I_{p}) ^ {\minus1} S U^{T} Y$.}
\label{ridgeCV vs MOR vs B-MOR}
\end{figure}

In brain encoding, ridge regression independently fits a regression model on each spatial target.  Therefore, ridge regression can be easily parallelized into multiple sub-models addressing different targets. For a given matrix $X$, scikit-learn's MultiOutputRegressor class subdivides the set of all target values $Y$ into $t$ sub-problems, each corresponding to a single spatial target. Ridge regression can now be expressed into solving $t$ independent estimations of the weights matrix $W$, as illustrated in Figure~\ref{ridgeCV vs MOR vs B-MOR}. This sub-division is repeated for each value of the regularization parameter $\lambda$. As all the targets and $\lambda$ values are independent, no communication is required between the sub-problems. Scikit-learn's MultiOutputRegressor class parallelizes the resolution of the sub-problems using a configurable number of concurrent processes $c$ executed with the \texttt{joblib} library. Joblib supports multiple execution backends including single-host thread-based or process-based parallelism, and distributed parallelism using the Dask~\cite{Dask} or Ray~\cite{moritz2018ray} engines.  We used the Dask backend and launched its distributed scheduler that simultaneously manages the computation requests and tracks the compute node  statuses.

\subsubsection{Proposed distributed ridge regression: Batch Multi-Output  Regression (B-MOR)} 

\label{sec:bmor}
This approach reduces the amount of redundant computations  by partitioning the problem into a  number of sub-problems equal to the number of available compute nodes in the distributed system, denoted as $c$ (Figure \ref{ridgeCV vs MOR vs B-MOR}). This strategy preserves maximal parallelism while reducing computational overheads. 
Algorithm~\ref{alg:b-mor} describes the approach. It consists of a main parallel for loop where the target output matrix ($Y$) is partitioned into $n$ sub-problems, where $n$ is the minimum value between the number of targets and the number of compute nodes. Each sub-problem represents a batch of targets $Y_1, \cdots, Y_{n}$. The algorithm uses the following helper functions:
\begin{itemize}
    \item \texttt{split} returns training and validation sets associated with a given cross-validation split.
    \item \texttt{svd} computes the singular-value decomposition of a matrix.
    \item \texttt{eval\_score} computes the regression performance score from predicted and true values. Higher scores denote better performance.
\end{itemize}

\begin{algorithm}
\SetNoFillComment
\SetKwInOut{Input}{input}
\SetKwInOut{Output}{output}
\SetKwFor{Parfor}{parfor}{do}{end}
\caption{Batch Multi-Output Regression (B-MOR)}
\label{alg:b-mor}
\Input{X---Input stimuli feature matrix}
\Input{Y---Target matrix}
\Input{t---Number of targets}
\Input{$\lambda$---Candidate hyper-parameters}
\Input{c---Number of concurrent jobs}

\Output{B---List of trained weight matrices for each sub-problem}

$n \leftarrow \min(t, c)$\;
\tcp{Main parallel for loop}
\Parfor{$i = 0$ \KwTo $n - 1$}{
    \tcp{Divide the target matrix Y into $n$ sub-problems}
    $Y_i \leftarrow$ Sub-matrix of $Y$ with columns $\left[\frac{i \cdot t}{n}, \frac{(i+1) \cdot t}{n}\right]$\;
    \For{all cross-validation splits $s$}{
        $X_{\text{train}}, X_{\text{val}}, Y_{\text{train}}, Y_{\text{val}} \leftarrow \text{split}(s, X, Y_i)$\;
        $USV^T \leftarrow \text{svd}(X_{\text{train}})$\;
        \For{all $\lambda$}{
            $M_{\lambda} \leftarrow V (S^2 + \lambda I_p)^{-1} S U^T$\;
            $\hat{Y}_{\text{val}} \leftarrow X_{\text{val}} M_{\lambda} Y_{\text{train}}$\;
            $\text{score}[i, s, \lambda] \leftarrow \text{eval\_score}(\hat{Y}_{\text{val}} ,  Y_{\text{val}})$\;
        }
    }
    \tcp{Calculate mean score across cross-validation splits}
    \For{all $\lambda$}{
        $\text{mean\_score}[i, \lambda] \leftarrow \frac{1}{|s|}\sum_{s} \text{score}[i, s, \lambda]$\;
    }
    \tcp{Find the best hyperparameter $\lambda$ for each sub-problem}
    ${\text{best}\_\lambda}[i] \leftarrow \arg\max_{\lambda} \{\text{mean\_score}[i, \lambda]\}$\;
    $B[i] \leftarrow M_{{\text{best}\_\lambda}[i]} Y_i$\;
}
\Return B\;

\end{algorithm}

\section{Complexity analysis}
\label{sec:complexity}
\subsection{Ridge regression with a single thread}
Ridge regression for a given hyper-parameter $\lambda$ is computed by:
$$ \hat{Y}_\lambda = XM_\lambda Y.$$

In this section, we outline the time complexity $T_M$ of computing $M_\lambda$ as well as the time complexity $T_W$ of computing the multiplications $XM_\lambda Y$. Matrix notations as well as their dimensions are summarized in Table \ref{tab:matrix_sizes}. Time complexities are expressed in the number of floating-point multiplications.

The cost $T_W$ of computing $XM_\lambda Y$ independently over $r$ values of hyper-parameter $\lambda$ is:
$$
T_W = O(pntr).
$$

Regarding $T_M$, Equation ~\ref{eq:M_computation}  requires inverting a square matrix of size $p$ ---time complexity $O(p^3)$--- and multiplying the resulting inverted matrix with matrix $X^T$ of size $p \times n$ ---time complexity O($p^2n$). Matrix $M_\lambda$ is computed once for all the targets. With a single hyper-parameter value (r=1), Equation~\ref{eq:M_computation} thus gives a complexity of $O(p^3 + p^2n)$.

If we were to naively iterate this approach for $r$ hyper-parameter values, the resulting time complexity would be $O(p^3r + p^2nr)$. However, expressing matrix $M_\lambda$ from the SVD of matrix $X$ as in Equation~\ref{eq:M_svd} reduces the time complexity of computing M to:
$$T_M = O(p^2nr+pr)$$

Indeed, computing the SVD of $X$ has time complexity $O(p^2n)$ since p $\leq$ n, the computation of $(S^2 + \lambda I_{p}) ^ {\minus1} SU^T$ has time complexity $O(p)$ since $S$ is diagonal, and the multiplication of $V$ with $(S^2 + \lambda I_{p}) ^ {\minus1} SU^T$ has time complexity $O(p^2n)$.

Finally, the overall time complexity $T_{\textrm{ridge}}$ of ridge regression iterated on $r$ hyper-parameter values, including computation of $M_\lambda$ and multiplication by matrix Y is:
$$T_{\textrm{ridge}} = T_M + T_W = O(p^2nr+pr+pntr)$$

\begin{table}[htbp]
    \centering
    \caption{Matrix Sizes. $n$: number of time samples; $p$: number of ANN features; $t$: number of brain targets. Other important notations include $c$: number of concurrent distributed executions and $r$: number of hyper-parameters.}
    \label{tab:matrix_sizes}
    \begin{tabular}{|c|c|c|}
    \hline
    Matrix & Dimensions & Description \\
    \hline
    $X$ & $n \times p$ & Feature matrix \\
    $Y$ & $n \times t$ & Brain target matrix \\
    $M_\lambda$ & $p \times n$ & Resolution matrix \\      
    $U$ & $n \times p$ & Left singular matrix \\
    $S$ & $p \times p$ & Singular values matrix \\
    $V$ & $p \times p$ & Right singular matrix \\
    \hline
    \end{tabular}
\end{table}

\subsection{Ridge regression with MOR}
In the case of MOR, the matrix multiplication $M_\lambda Y$ is replaced by $t$ multiplications of $M$ with a vector $y$, which does not change the time complexity. Provided that the number of targets $t$ is larger than the number of concurrent computing nodes $c$ ---which is the case in our application---, all these matrix-vector multiplications happen in parallel, and the resulting time complexity on the application critical path is $c^{-1}T_W$. 

By contrast, the computation of matrix $M_\lambda$ is repeated independently for each brain target, resulting in a massive overhead computation $tT_M$ distributed over $c$ concurrent processes. Overall the computational cost of ridge regression implemented with MOR is:
\begin{equation}
\label{Eq:TMOR}
T_{\textrm{MOR}} = c^{-1}(T_W + tT_M).
\end{equation}

\subsection{Ridge regression with B-MOR}
B-MOR scales better than the previous approach due to the use of $c$ sub-problems instead of $t$. The computation of the matrix multiplication costs $c^{-1}T_W$, similar to MOR. However, the overhead of recomputing matrix $M_\lambda$ for each sub-problem is only $cT_M$, which is distributed across $c$ concurrent execution. The computational cost of ridge implemented with B-MOR is thus:
\begin{equation}
\label{Eq:TBMOR}
T_\textrm{B-MOR} = c^{-1}T_W + T_M.
\end{equation}

Comparing Equations \ref{Eq:TMOR} and \ref{Eq:TBMOR}, we observe that the time complexity of the B-MOR implementation is much lower than for the MOR implementation, as $T_{MOR}-T_{B-MOR}=(c^{-1}t-1)T_M$. This difference may be massive when $c<<t$. We also observe that when $c>1$, B-MOR has lower time complexity than single-threaded ridge regression. However, the parallel efficiency of B-MOR is limited by the term $T_M$ which is not reduced by $c$. 

\section{Results}

We report on a series of benchmark experiments for brain encoding, using scikit-learn's multithreaded, MOR and B-MOR implementations of ridge regression with hyper-parameter optimization across 11 values of $\lambda$ . The benchmarks were applied to the Friends CNeuroMod dataset (N=6 subjects) at multiple spatial resolutions to investigate the scalability of different implementations of ridge regression (parcel-wise, ROI-wise, and truncated versions of whole-brain voxel-wise time series).

\subsection{Brain encoding models successfully captured brain activity in the visual cortex}
\begin{figure}[htbp]
%\hspace* {-2 cm}
\centering
\includegraphics[width=14cm]{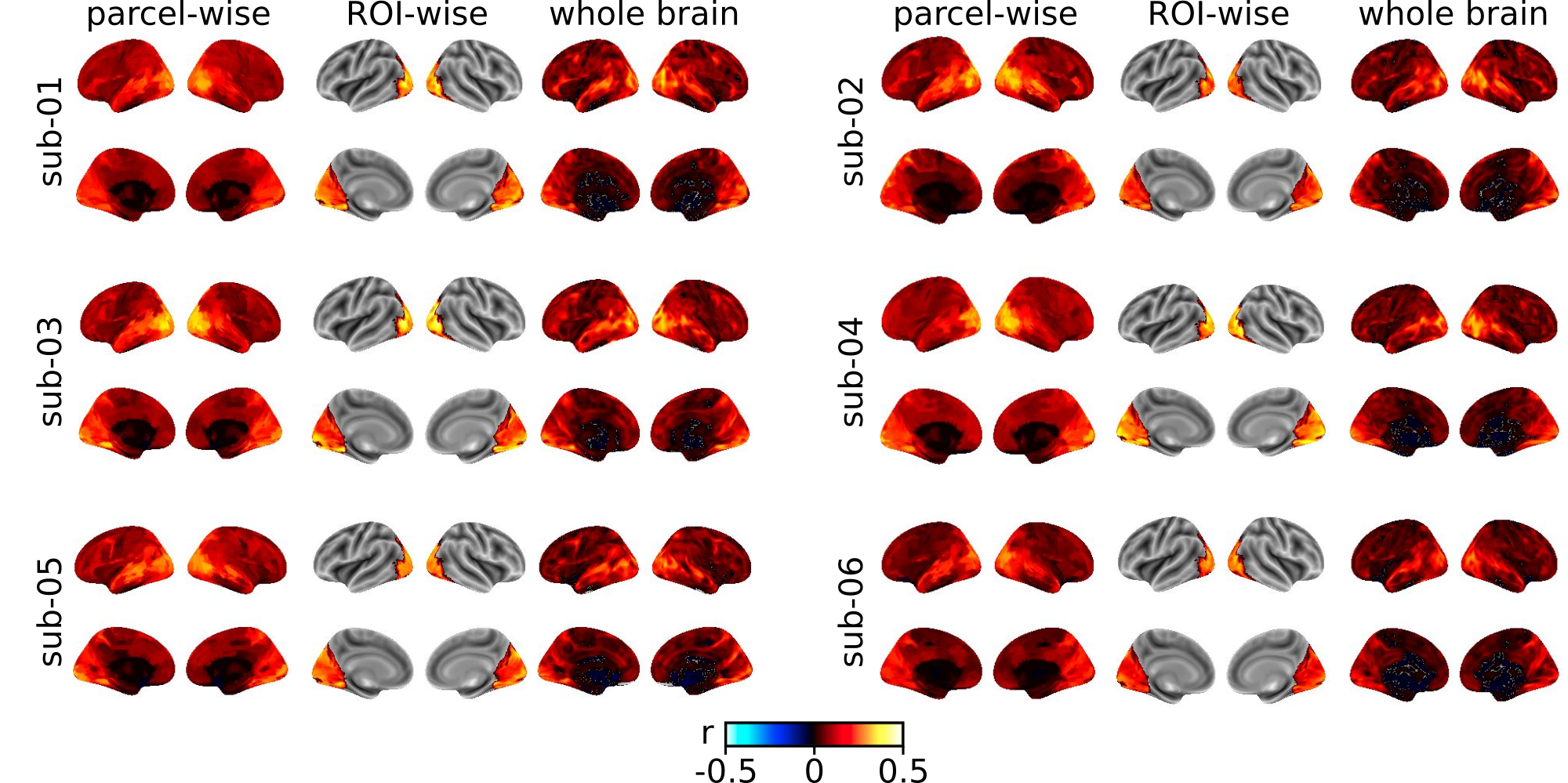}
\caption{Brain encoding results, with performance based on Pearson Correlation Coefficient (r) between real and predicted time series in the friends dataset (N=6 subjects).}
        \label{Results_figure_map_n}
\end{figure}

We first aimed to validate that our brain encoding model performed in line with recent studies, at the individual level and for different resolutions of target brain data. 

We extracted features of dynamic visual stimuli, by selecting a subset of images in the video, and feeding these images independently into an established vision pretrained network called VGG16 \cite{VGG16_ref}. We used 3 seasons of Friends TV show, where $10\%$ of data was set aside as the test data. Brain encoding was implemented using ridge regression, and the best value of hyper-parameter $\lambda$ was selected through cross-validation. 

Figure~\ref{Results_figure_map_n} shows the functional alignment between the features of the last fully connected layer (FC2) of VGG16  brain activities and brain activity for six subjects. Brain encoding maps were highly consistent across subjects and resolutions of analysis. In all cases, a moderate correlation, up to 0.5, was observed in the visual cortex between the real fMRI time series and the time series predicted by the brain encoding model. For the analysis that included the full brain, moderate accuracy for brain encoding was observed in other brain areas as well, such as temporal cortices involved in high level visual processing as well as language. Analysis at the voxel level brought more anatomical details but were overall consistent with brain encoding maps at the parcel level. 

Overall, brain encoding models successfully predicted activity in expected brain regions, for all subjects and resolutions. 

\subsection{Brain encoding was significant compared to a null distribution}

\begin{figure}[!ht]
    \begin{subfigure}[b]{0.49\textwidth}
        \centering
        \includegraphics[width=1\linewidth]{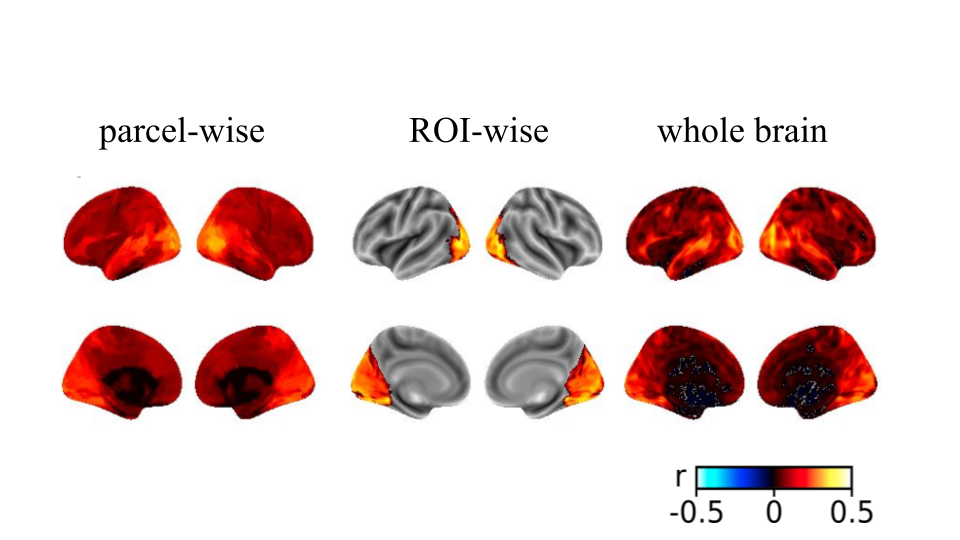}
  \caption{Trained brain encoding}
    \end{subfigure}
        \begin{subfigure}[b]{0.49\textwidth}
        \centering
        \includegraphics[width=1\linewidth]{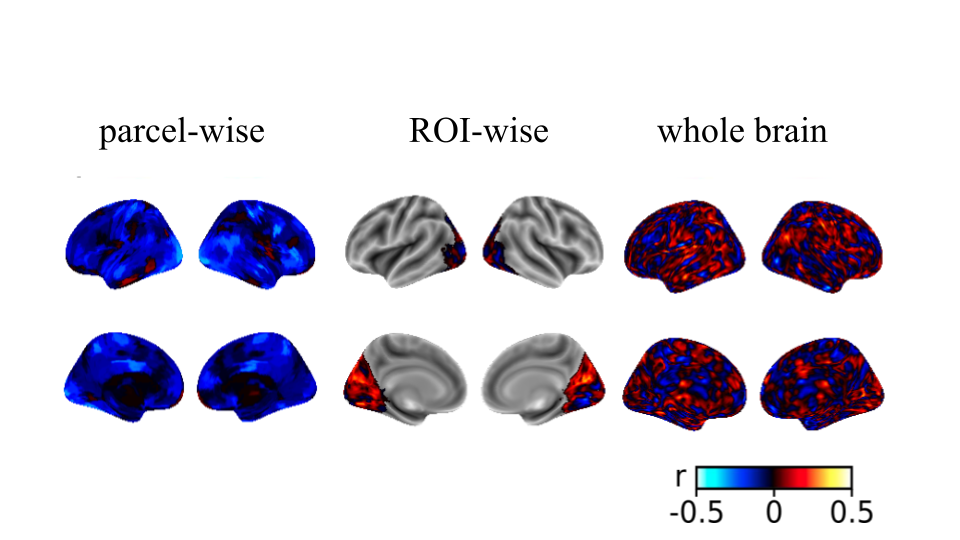}
        \caption{Untrained brain encoding} 
    \end{subfigure}
    \caption{Brain encoding predictions for a single individual (sub-01) in two cases. Panel \textbf{a}: corresponding pairs of \{fMRI time series and stimuli\} were presented to the ridge regression models. Panel \textbf{b}: random permutations of fMRI time series and stimuli data were presented to the ridge regression model.}
    \label{fig:comparison_null} 
\end{figure}

Next, we wanted to assess the significance of the brain encoding, compared to a null distribution were the movie frames used as input to the model did not correspond with brain data time series. We repeated the brain encoding procedure presented in the previous section for one subject (sub-01), after random shuffling of the image features and brain images. Figure \ref{fig:comparison_null} compares the original brain encoding results (panel \textbf{a}) with brain encoding based on shuffled features (panel \textbf{b}). While the original brain encoding results reached moderate accuracy, up to 0.5 correlation between real and predicted brain activity, the performance using shuffled features was dramatically lower. The correlation values were typically an order of magnitude smaller, less than 0.05. The quality of brain encoding using original image frames thus appeared as significant compared to a null distribution where image features were randomly shuffled.

\begin{figure}[!ht]
    \begin{subfigure}[b]{0.475\textwidth}
        \centering
        \includegraphics[width=0.48\linewidth]{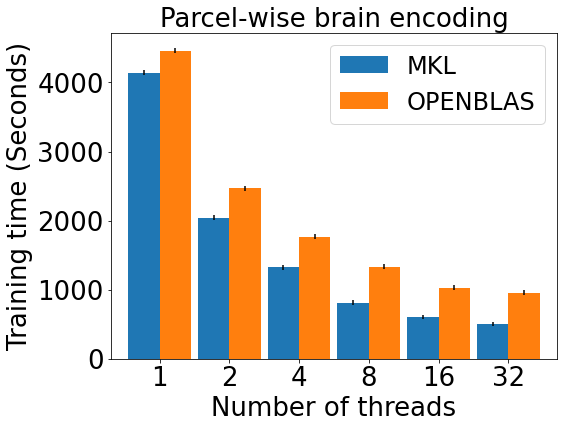}
        \includegraphics[width=0.48\linewidth]{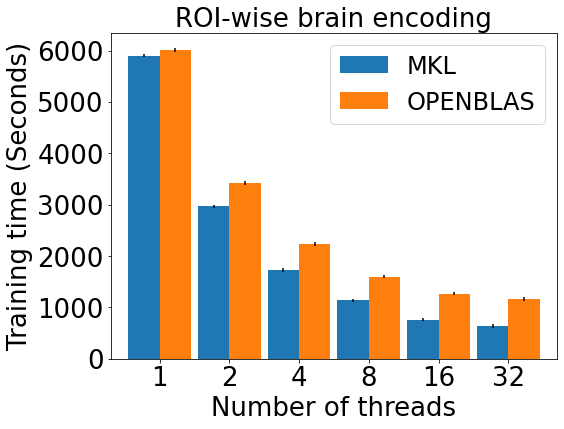}
        \caption{Sub-01}
    \end{subfigure}
    \begin{subfigure}[b]{0.475\textwidth}
    \centering
        \includegraphics[width=0.48\linewidth]{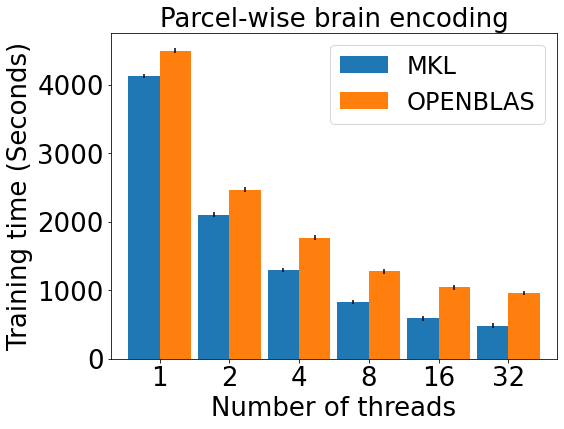}
        \includegraphics[width=0.48\linewidth]{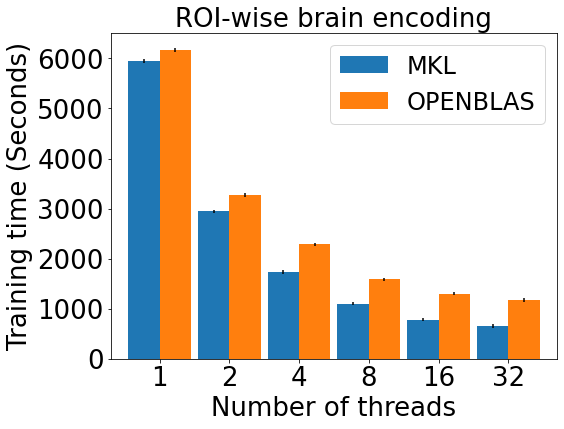}
        \caption{Sub-02}
    \end{subfigure}

    \begin{subfigure}[b]{0.475\textwidth}
        \centering
        \includegraphics[width=0.48\linewidth]{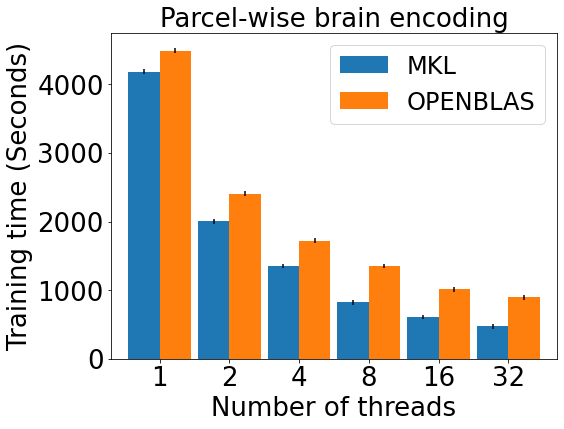}
        \includegraphics[width=0.48\linewidth]{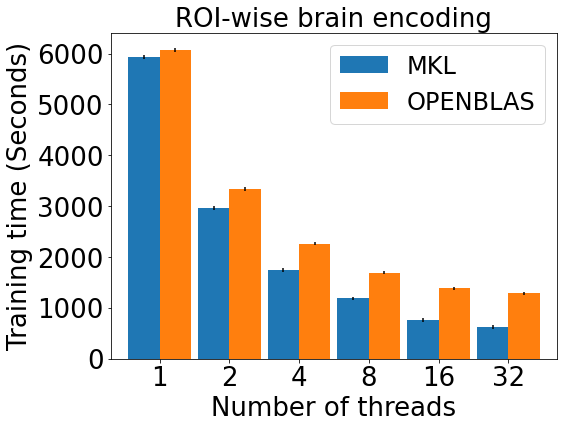}
        \caption{Sub-03}
    \end{subfigure} 
    \begin{subfigure}[b]{0.475\textwidth}
        \centering
        \includegraphics[width=0.48\linewidth]{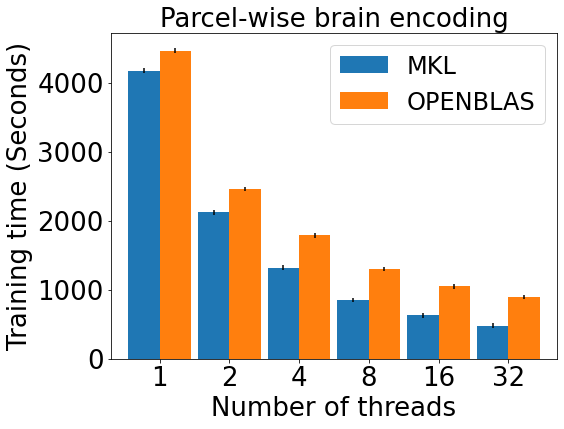}
        \includegraphics[width=0.48\linewidth]{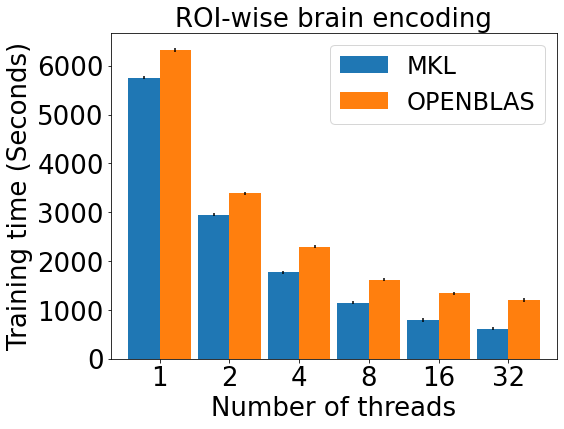}
        \caption{Sub-04}
    \end{subfigure}  

    \begin{subfigure}[b]{0.475\textwidth}
        \centering
        \includegraphics[width=0.48\linewidth]{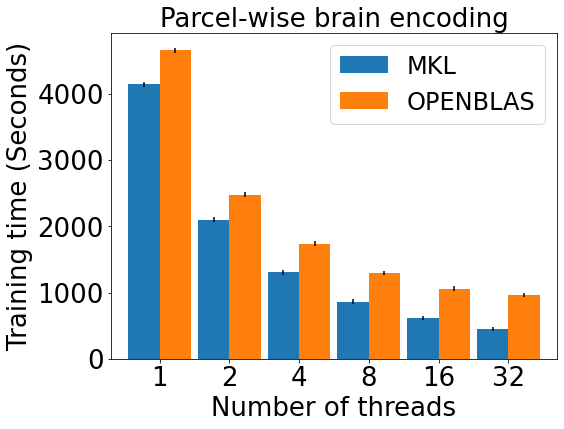}
        \includegraphics[width=0.48\linewidth]{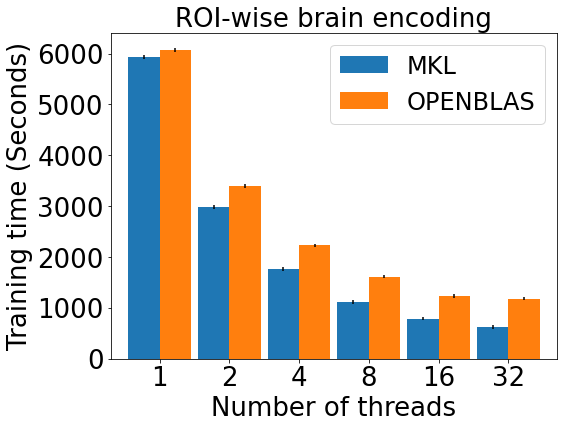}
        \caption{Sub-05}
    \end{subfigure}
    \begin{subfigure}[b]{0.475\textwidth}
        \centering
        \includegraphics[width=0.48\linewidth]{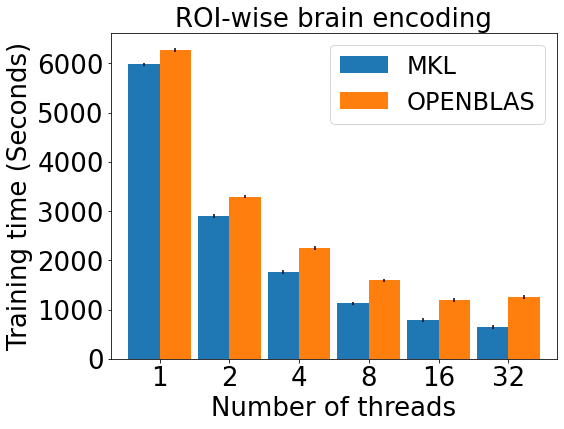}
        \includegraphics[width=0.48\linewidth]{Figure_6/Sub1_Pa.png}
        \caption{Sub-06}
    \end{subfigure}
  
    \caption{Comparison of MKL and OpenBLAS implementations for multithreaded execution.}
   \label{MKLvsOpenblas6subjects} 
\end{figure}

\subsection{Multithreaded execution with Intel MKL provided significant speedup compared to OpenBLAS }
After establishing the quality of our multiresolution brain encoding benchmark, we proceeded to compare the performance and scalability of ridge regression using scikit-learn on a 32-core compute node, and comparing the libraries underlying multithreaded parallelization, i.e. MKL and OpenBLAS. Figure~\ref{MKLvsOpenblas6subjects} illustrates the comparison between OpenBLAS  and MKL multi-threading for two different resolutions (parcel-wise and ROI-wise), six subjects, and varying numbers of parallel threads. The experiments with whole-brain resolution could not be completed due to out-of-memory limitation with our benchmark system. The results consistently demonstrated that the MKL library outperformed the OpenBLAS library for all subjects and thread configurations. Specifically, when using 32 threads, the MKL library exhibited a speedup factor of 1.90 and 1.98 compared OpenBLAS for parcel-wise and ROI, respectively, on average across all subjects. This indicates a substantial improvement in processing time with the MKL library compared to OpenBLAS.

\begin{figure}
    \begin{subfigure}[b]{0.49\textwidth}
        \centering
        \includegraphics[width=1\linewidth]{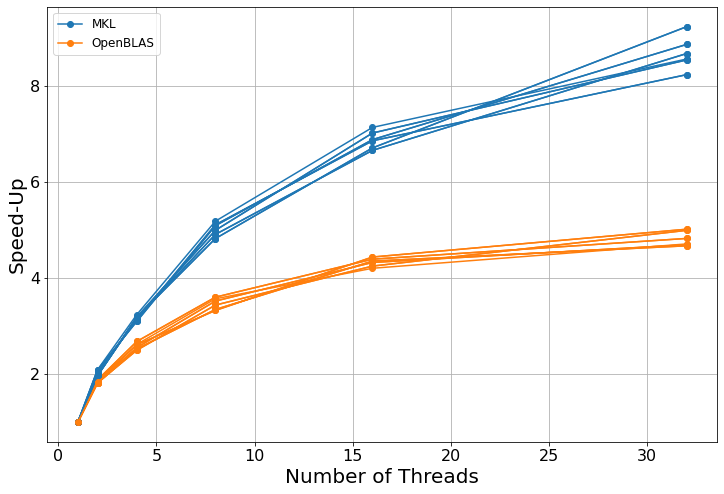}
         \caption{Parcel-wise}
    \end{subfigure}
    %\vskip\baselineskip
    \begin{subfigure}[b]{0.49\textwidth}
            \includegraphics[width=1\linewidth]{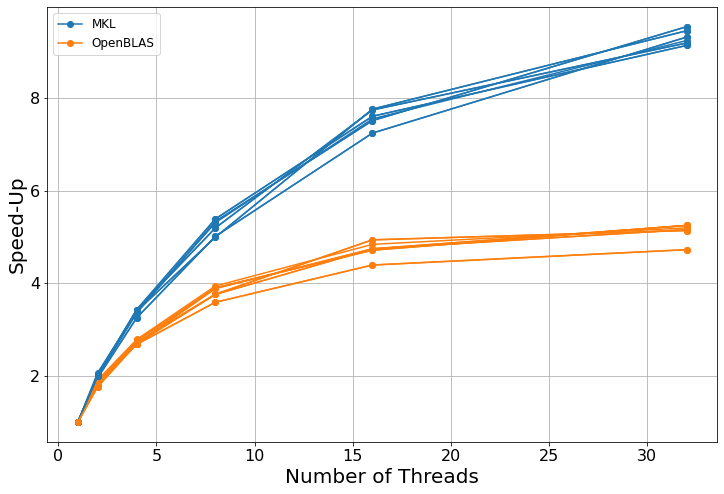}
              \caption{ROI-wise}
    \end{subfigure}
      
    \caption{Speed-up rates of ROI-wise ridgeCV execution time for MKL and OpenBLAS  across different number of threads. Each line on the plot corresponds to a specific subject. For each subject, the solid lines represent MKL, and the dashed lines represent OpenBLAS.}
   \label{Speedup1} 
\end{figure}

\subsection{Speed-up of multi-threading quickly reached a plateau with an increasing number of threads}
We also observed a sharp decrease in the efficiency of parallelization with an increasing number of threads. Figure \ref{Speedup1} represents the speed-up performance of two libraries MKL and OpenBLAS for parcel-wise and ROI-wise brain encoding across different numbers of threads. The parallelization speed-up is calculated as follows:
\[
SU = \frac{T_R}{T_P}
\]
where $SU$ is the speed-up, $T_R$ is the execution time with 1 thread, and $T_P$ is the execution time with 2, 4, 8, 16, or 32 threads. The speed-up measures how effectively the parallel resources are being used.

A consistent observation across subjects was that, as the number of threads increased, the parallel efficiency decreased, as expected according to Amdhal's law. In other words, as more threads were employed, the incremental improvement in speed-up rate was reduced. These diminishing returns in speed-up highlight the need for careful selection of thread count to balance computational resources with performance. 

\begin{figure}[!ht]
    \begin{subfigure}[b]{0.49\textwidth}
        \centering
        \includegraphics[width=1\linewidth]{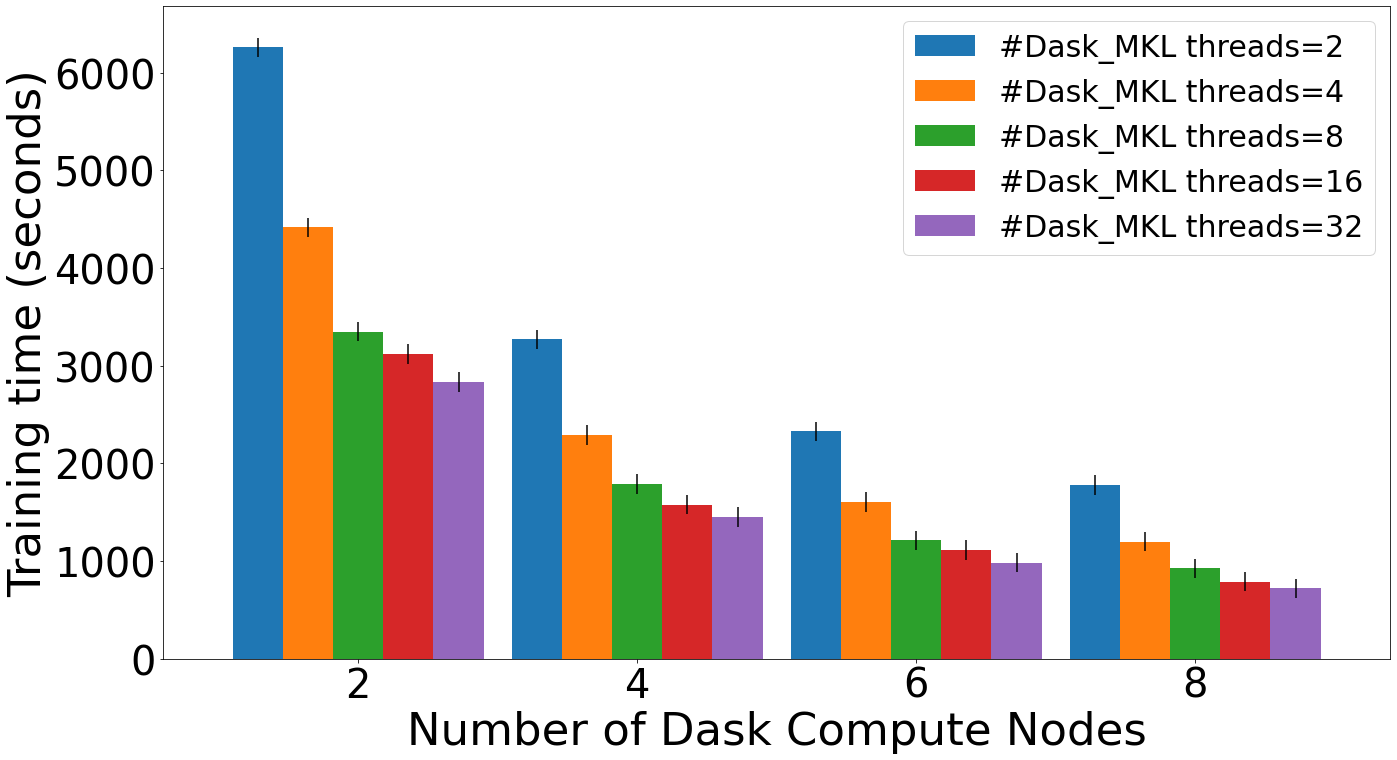}
         \caption{Sub-01}
    \end{subfigure}
    %\vskip\baselineskip
    \begin{subfigure}[b]{0.49\textwidth}
            \includegraphics[width=1\linewidth]{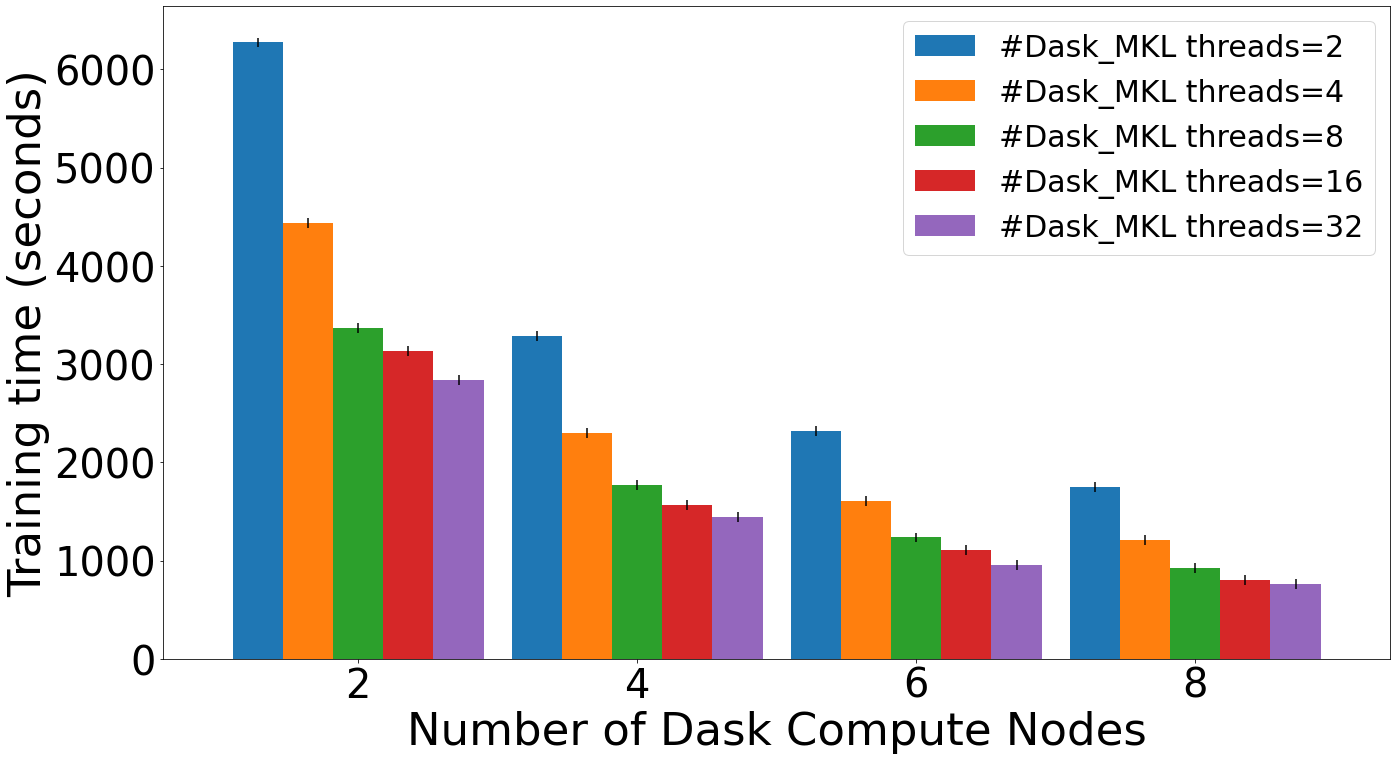}
              \caption{Sub-02}
    \end{subfigure}
    \\
    \begin{subfigure}[b]{0.49\textwidth}
        \centering
        \includegraphics[width=1\linewidth]{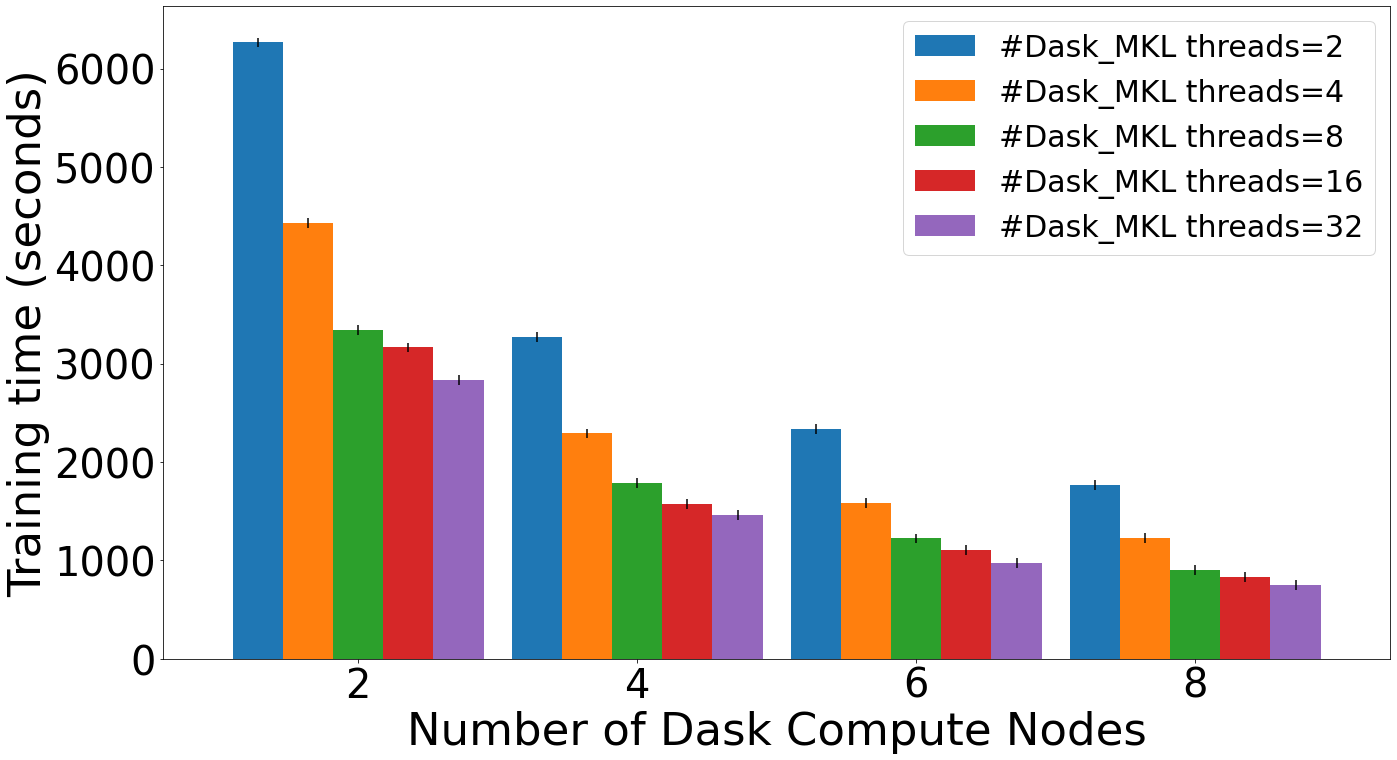}
               \caption{Sub-03}
    \end{subfigure} 
        \begin{subfigure}[b]{0.49\textwidth}
        \centering
        \includegraphics[width=1\linewidth]{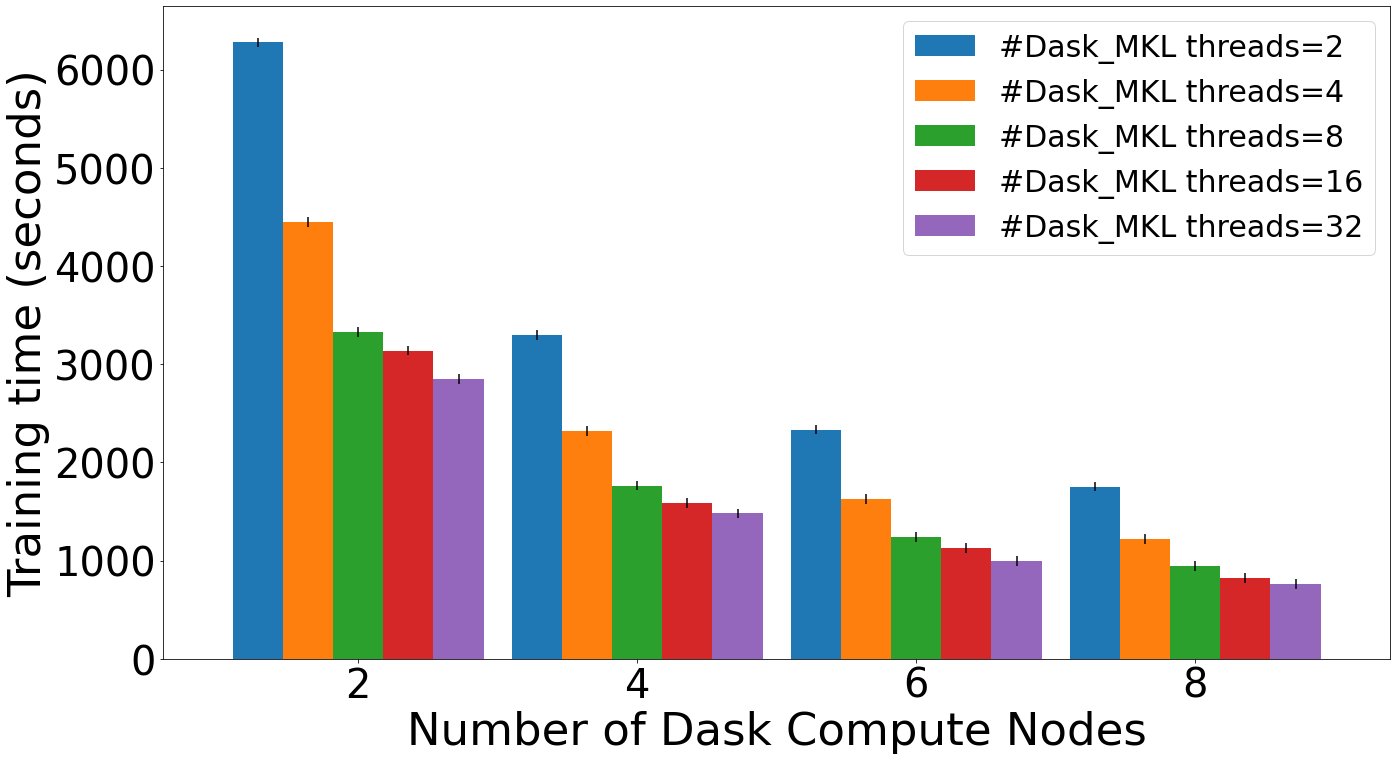}
                \caption{Sub-04}
    \end{subfigure}
    \\
    \begin{subfigure}[b]{0.49\textwidth}
        \centering
        \includegraphics[width=1\linewidth]{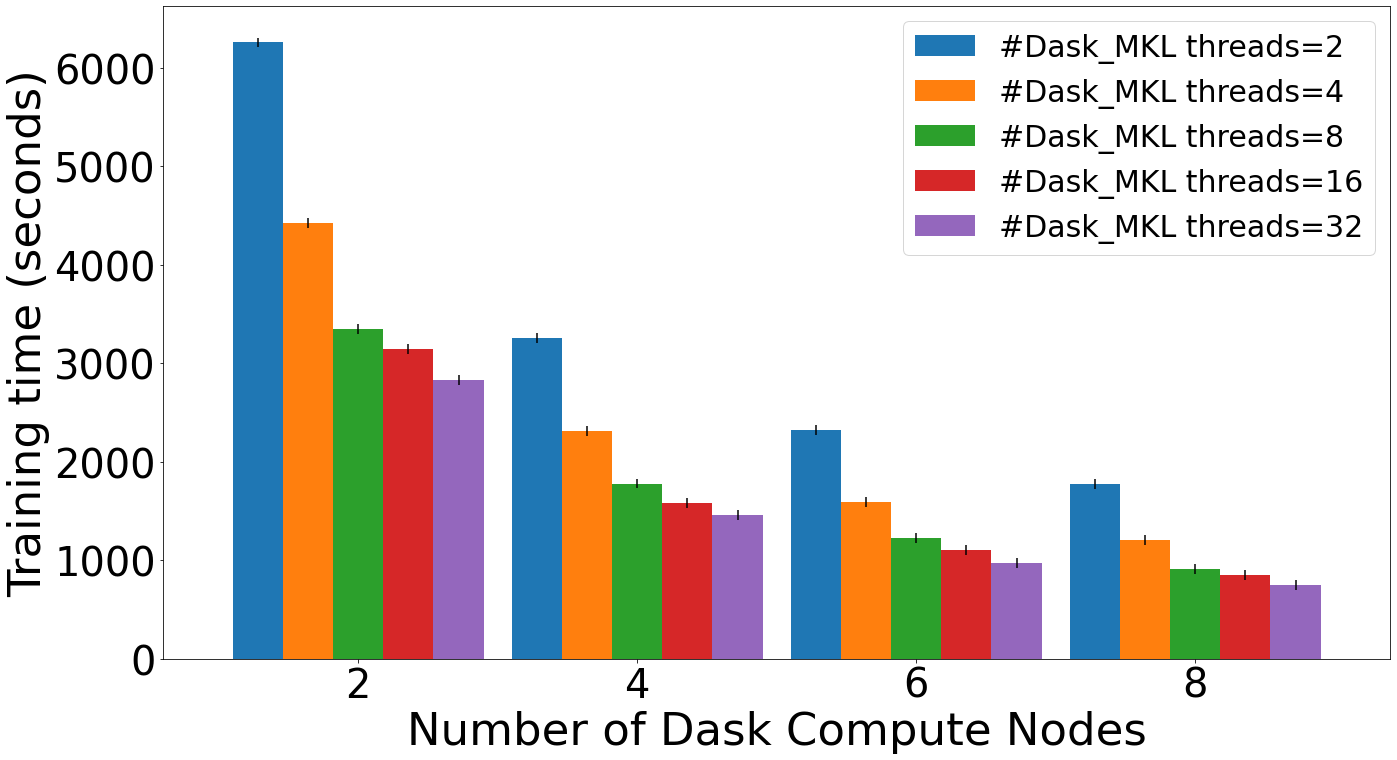}
                \caption{Sub-05}
    \end{subfigure}
    \begin{subfigure}[b]{0.49\textwidth}
        \centering
        \includegraphics[width=1\linewidth]{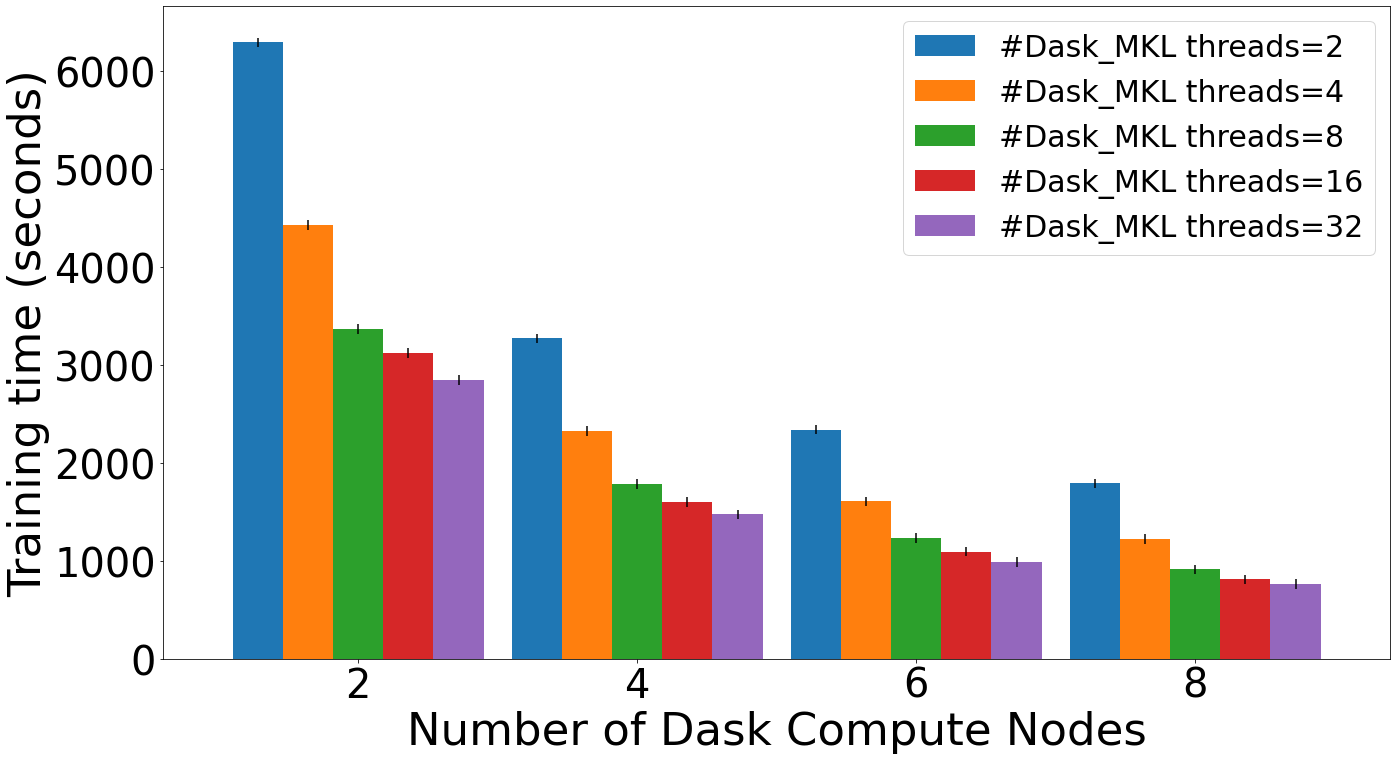}
                        \caption{Sub-06}
    \end{subfigure}
  
    \caption{MultiOutput ridgeCV training time for 6 subjects using whole brain (MOR) data described in Table~\ref{Tab:Tcr}. The MOR implementation scales across threads and Dask compute nodes, however, it has a massive overhead: multi-threaded scikit-learn implementation  with a single compute node and 32 threads takes approximately 1s.}
   \label{fig:MOR1} 
\end{figure}
\subsection{MultiOutput ridge regression scales across compute nodes and threads, but is much slower than multi-threading with RidgeCV}

In this next experiment, we implemented scikit-learn's original MultiOutput ridge regression (MOR) within a Dask-based distributed parallelism setting for brain encoding tasks. We chose to focus on whole-brain resolution for this experiment, as lower resolutions failed to show the advantage of parallelization with MOR. However, whole-brain resolution was too slow to run as is with MultiOutput ridge regression. Therefore, we created a custom truncated version of whole-brain resolution, called whole-brain (MOR), as seen in Table~\ref{Tab:Tcr}. Figure~\ref{fig:MOR1} reports on the parallelization of MultiOutput across six subjects using this dataset, where the training process was distributed across multiple compute nodes and threads.

Figure \ref{fig:MOR1} shows a substantial reduction in training time with an increasing number of threads and compute nodes, for all subjects, which illustrates the good scalability of MOR parallelization. However, compute time was massively increased compared to the multi-threaded scikit-learn implementation on a single compute node. For example, using 8 compute nodes and 32 threads, compute time with MOR is in the order of 1000s, whereas the multi-threaded scikit-learn implementation with a single compute node and 32 threads takes approximately 1s. This overhead directly results from the increase in time complexity reflected in Equation~\ref{Eq:TMOR}. 

\begin{figure}[!ht]
    \begin{subfigure}[b]{0.49\textwidth}
        \centering
        \includegraphics[width=1\linewidth]{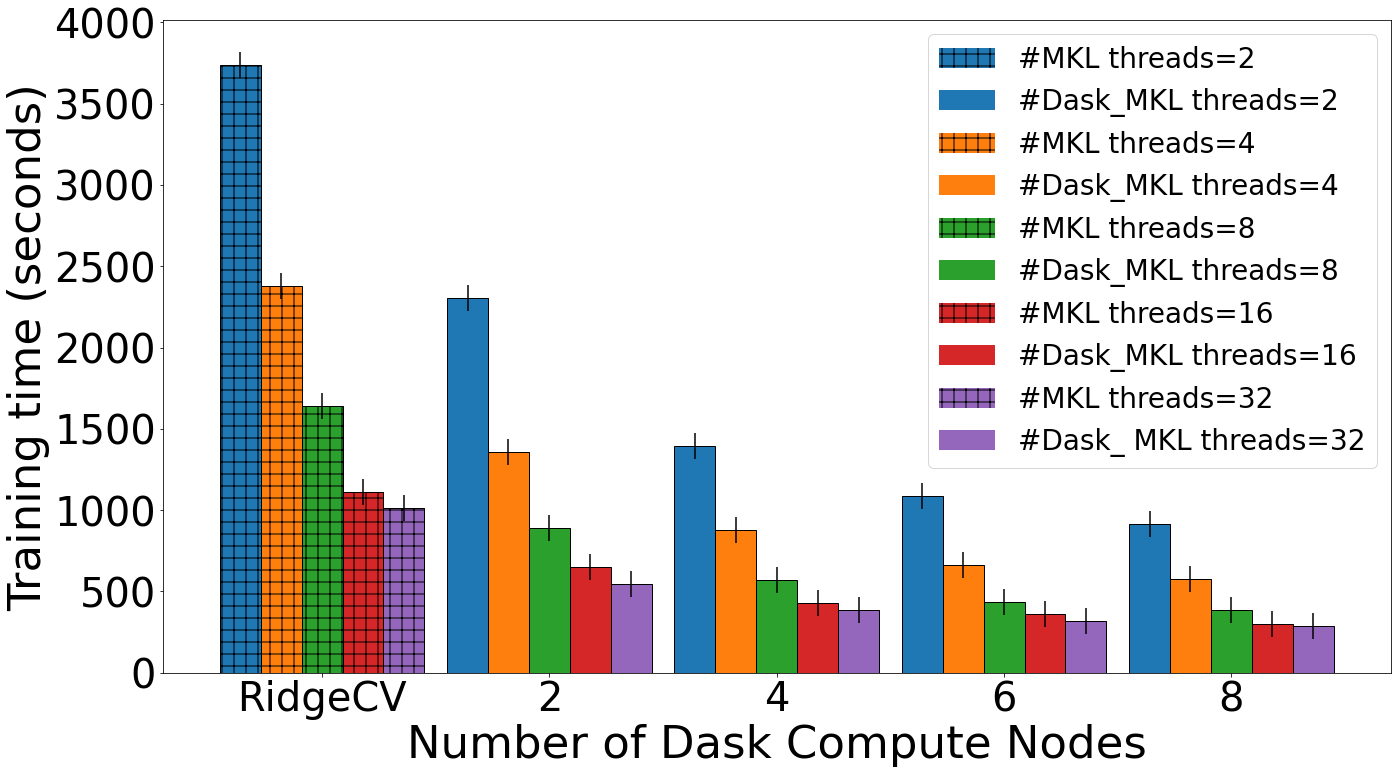}
         \caption{Sub-01}
    \end{subfigure}
    %\vskip\baselineskip
    \begin{subfigure}[b]{0.49\textwidth}
            \includegraphics[width=1\linewidth]{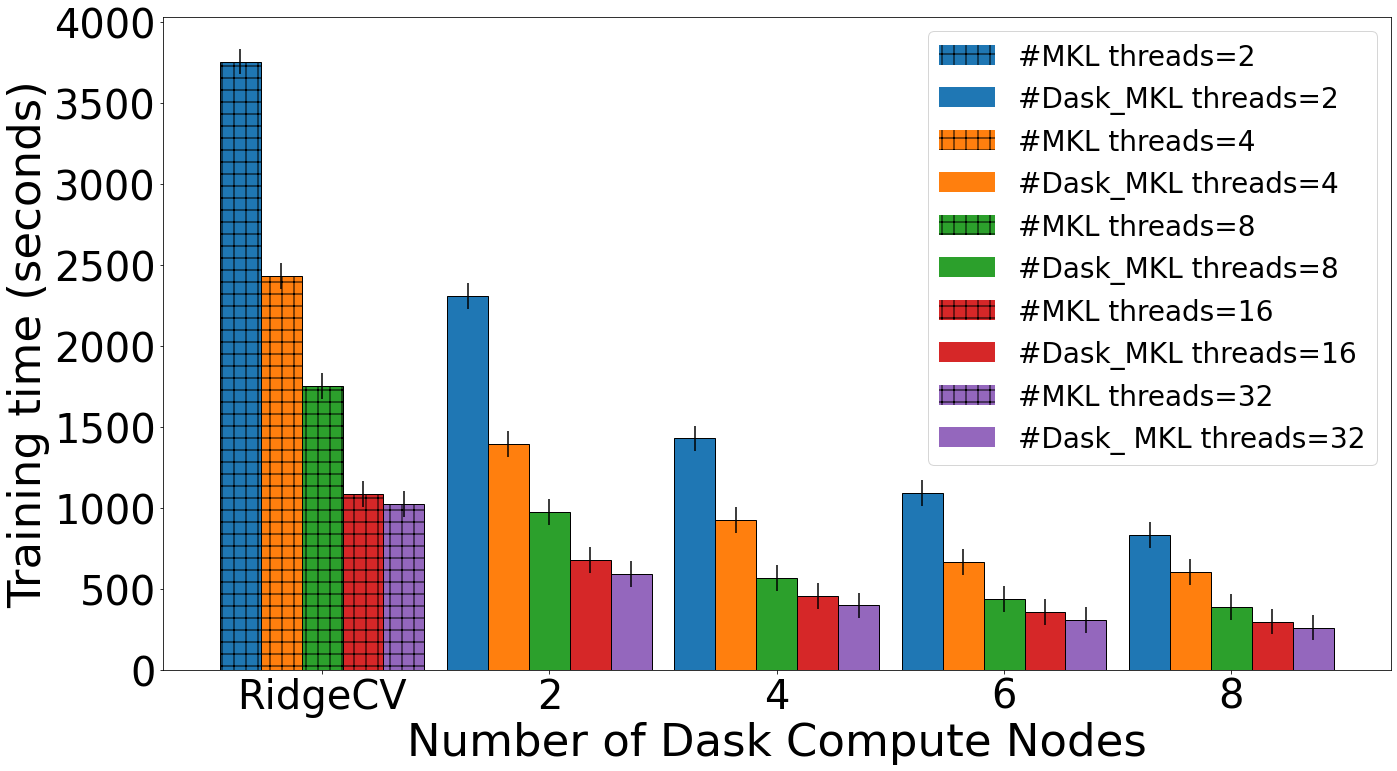}
              \caption{Sub-02}
    \end{subfigure}
    \\
    \begin{subfigure}[b]{0.49\textwidth}
        \centering
        \includegraphics[width=1\linewidth]{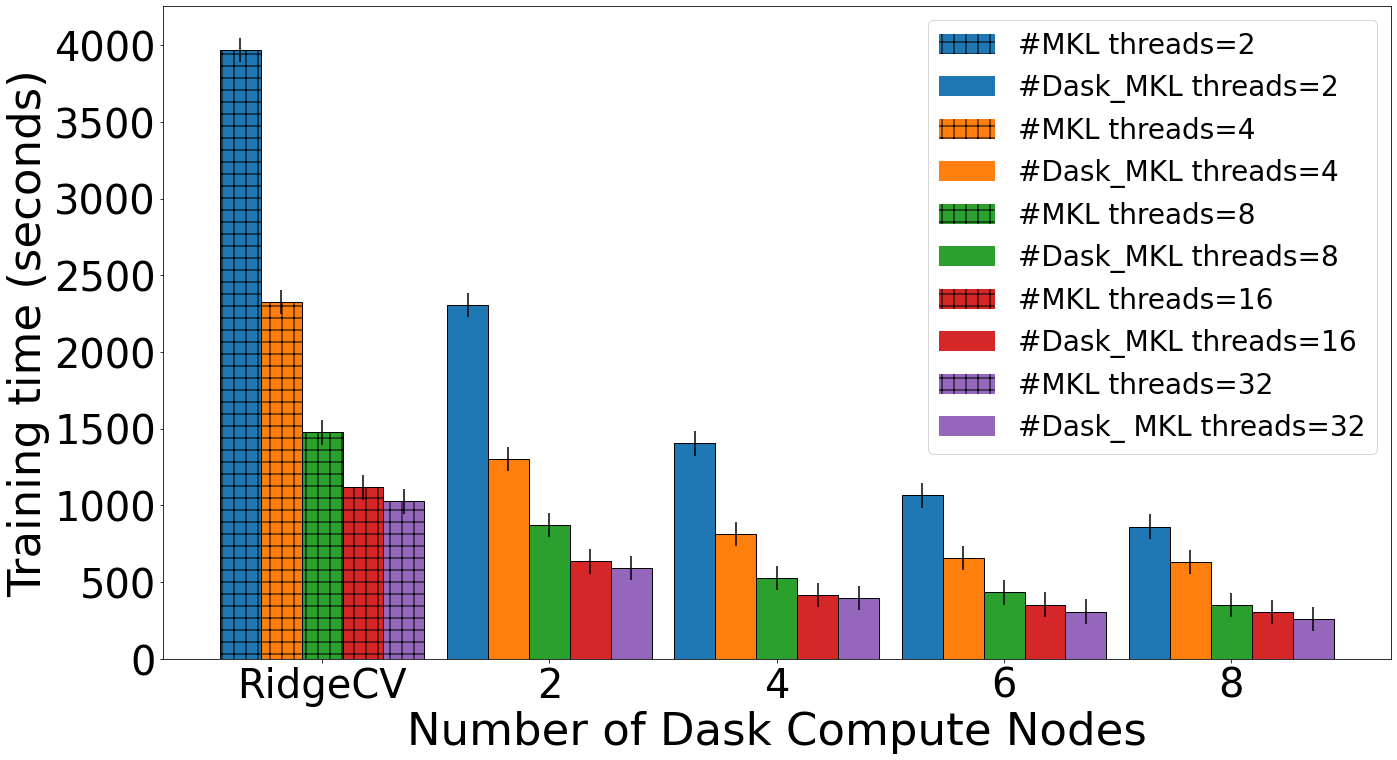}
               \caption{Sub-03}
    \end{subfigure} 
        \begin{subfigure}[b]{0.49\textwidth}
        \centering
        \includegraphics[width=1\linewidth]{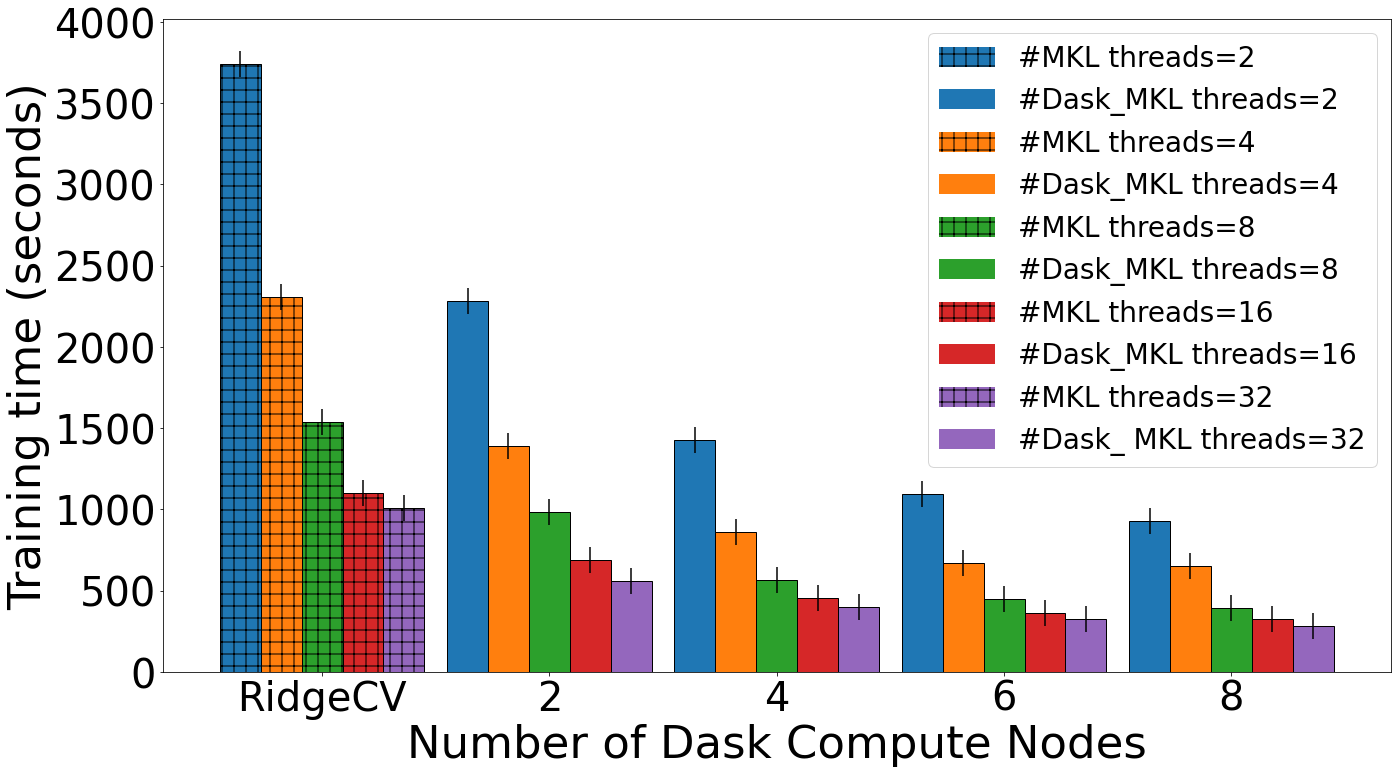}
                \caption{Sub-04}
    \end{subfigure}
    \\
    \begin{subfigure}[b]{0.49\textwidth}
        \centering
        \includegraphics[width=1\linewidth]{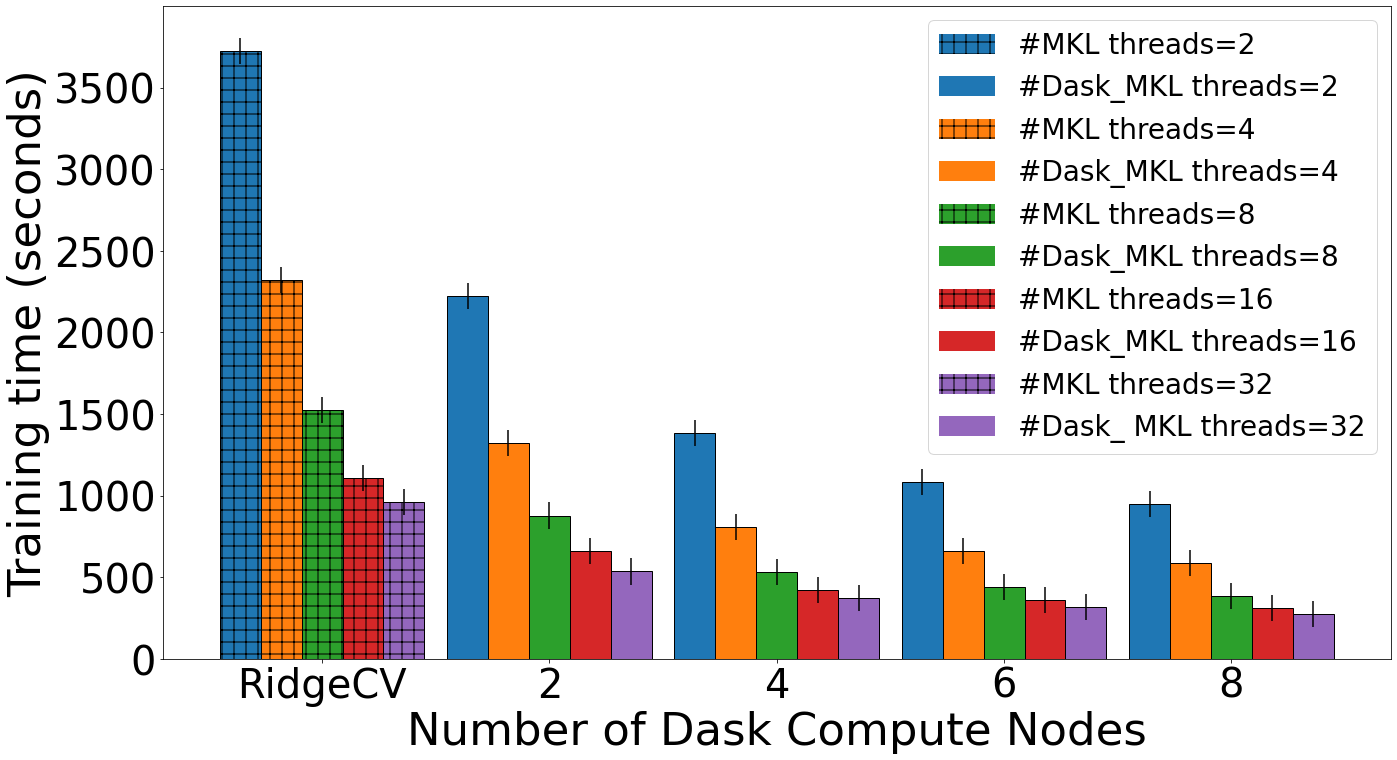}
                \caption{Sub-05}
    \end{subfigure}
    \begin{subfigure}[b]{0.49\textwidth}
        \centering
        \includegraphics[width=1\linewidth]{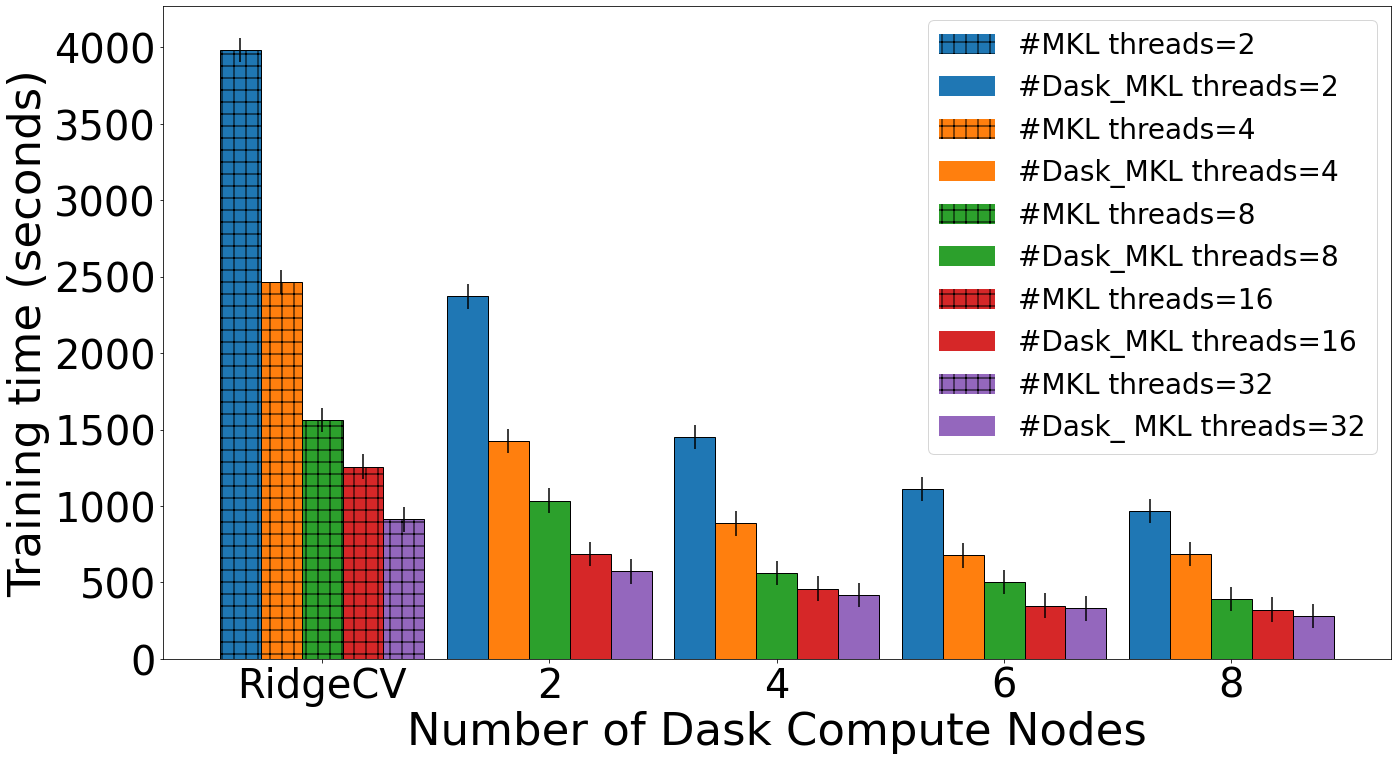}
                        \caption{Sub-06}
    \end{subfigure}
  
    \caption{B-MOR ridgeCV training time  6 subjects with whole brain (B-MOR) data described in Table~\ref{Tab:Tcr}. B-MOR scales across compute nodes and threads, and provides substantial speed-up compared to scikit-learn's multi-threaded implementation (labelled as ``RidgeCV").}
   \label{fig:MOR2} 
\end{figure}

\subsection{Batch multi-output regression leads to efficient speed-up across multiple compute nodes and threads}

\begin{figure}[!ht]
    \begin{subfigure}[b]{0.49\textwidth}
    \centering
    \includegraphics[width=1\linewidth]{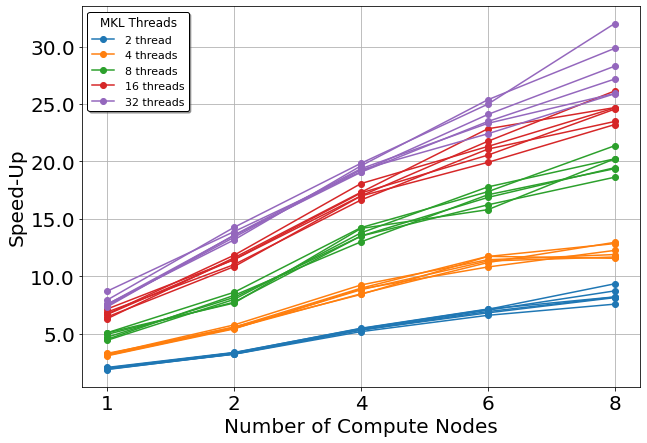}
    
    \caption{Speed-Up vs Number of Threads}
    \end{subfigure}    
    \begin{subfigure}[b]{0.49\textwidth}
    \centering
    \includegraphics[width=1\linewidth]{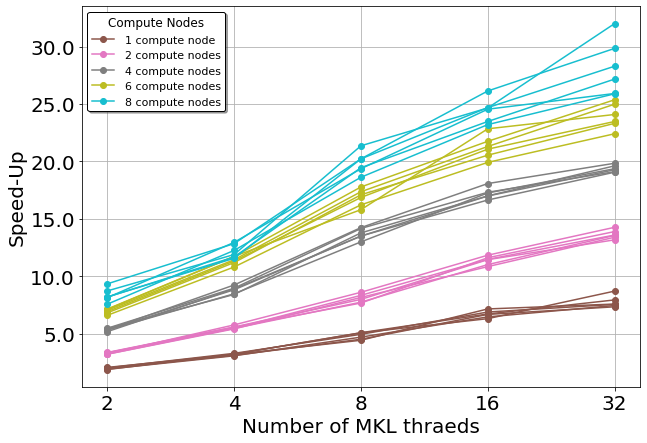}
    
    \caption{Speed-Up vs Number of Compute Nodes}
    \end{subfigure}
     
        \caption{Speed up in B-MOR training time for truncated B-MOR data across 6 subjects with varying numbers of threads and compute nodes in the Dask distributed system.}
    \label{speedup2} 
\end{figure}

In the next experiment, we benchmarked our B-MOR implementation of ridgeCV, that divides the brain targets into batches, and runs scikit-learn's multi-threaded RidgeCV on each batch independently with different compute nodes. Figure~\ref{fig:MOR2} shows that, as the number of threads and compute nodes increased, substantial speed-up in training time was achieved compared to scikit-learn's multithreaded implementation (labelled ``RidgeCV" in the figure), which demonstrates the practical value of the B-MOR implementation. To quantify this observation, we computed the distributed speed-up ratio as follows:
\[
\textrm{DSU} = \frac{T_R}{T_P}
\]
where $T_R$ indicates the execution time of scikit-learn original ridge regression on a single compute node and 1 thread, and $T_P$ indicates computation time with B-MOR for a given number of compute nodes and threads. Overall, the distributed speed-up ratio increased as the number of threads or compute nodes increased (Figure~\ref{speedup2}). The training time for B-MOR was approximately $30-33$ times less than the original scikit-learn ridge regression with 1 thread and 1 compute node. As it was the case with the multi-threaded implementation, the DSU reached a plateau beyond a certain number of compute nodes and threads, with diminishing performance returns as parallelization overheads and the time spent in unparallelized code sections started to outweigh parallelization benefits.

\section{Conclusion}
In this paper, we evaluated the efficiency of different implementations of ridge regression for a specific application: brain encoding using a vision model (VGG16) during movie watching. We found that the multithreaded parallelization available in scikit-learn could be used to reduce substantially computation time, and that the BLAS implementation provided by  the proprietary Intel oneAPI Math Kernel Library (MKL) substantially outperforms the open-source OpenBLAS implementation. For increased scalability, the Dask-based scikit-learn MultiOutput implementation can parallelize computations across multiple compute nodes, but this comes with massive redundancy in some of the computations, which we found to be impractical when using high-resolution brain targets (tens to hundreds of thousands). Therefore, we implemented a more efficient version of the MultiOutput method (B-MOR), that parallelizes ridge regression across batches of brain targets. Our B-MOR method scales well, both in terms of the number of compute nodes and the number of threads used by nodes. This approach allowed us to generate brain encoding maps with high spatial resolution and whithin a reasonable time. Our method could be useful for fMRI researchers who want to process high-resolution deep datasets with high-performance computing clusters. Our conclusion likely applies to any ridge regression for data arrays with a very large number of targets (up to the order of 100k) and a large number of predictors (in the order of thousands). As our proposed method is straightforward to implement, it may become available in scikit-learn in the future.

\section{Availability of code and data}

The code to reproduce our experiments is available at \url{https://github.com/Sana3883/Scaling-up-Ridge}. The CNeuroMod dataset is available at \url{https://www.cneuromod.ca/gallery/datasets}.

\section{Acknowledgments}
The computing platform used in the experiments was obtained with funding from the Canada Foundation for Innovation. The Courtois project on neural modelling was made possible by a generous donation from the Courtois foundation, administered by the Fondation Institut Gériatrie Montréal at CIUSSS du Centre-Sud-de-l’île-de-Montréal and University of Montreal. The Courtois NeuroMod team is based at “Centre de Recherche de l’Institut Universitaire de Gériatrie de Montréal”, with several other institutions involved. See the cneuromod documentation for an up-to-date list of contributors (https://docs.cneuromod.ca). PB is a senior fellow (chercheur boursier) from Fonds de Recherche du Québec (Santé). 

\bibliographystyle{model1-num-names}
\bibliography{sample.bib}

\begin{thebibliography}{42}
\expandafter\ifx\csname natexlab\endcsname\relax\def\natexlab#1{#1}\fi
\providecommand{\bibinfo}[2]{#2}
\ifx\xfnm\relax \def\xfnm[#1]{\unskip,\space#1}\fi
%Type = Article
\bibitem[{Naselaris et~al.(2011)Naselaris, Kay, Nishimoto, and Gallant}]{Encoding1}
\bibinfo{author}{T.~Naselaris}, \bibinfo{author}{K.~Kay}, \bibinfo{author}{S.~Nishimoto}, \bibinfo{author}{J.~Gallant},
\newblock \bibinfo{title}{Encoding and decoding in fmri. neuroimage},
\newblock \bibinfo{journal}{Technometrics} \bibinfo{volume}{56} (\bibinfo{year}{2011}) \bibinfo{pages}{400--410}.
%Type = Article
\bibitem[{Hoerl and Kennard(1970)}]{ridge1}
\bibinfo{author}{A.~Hoerl}, \bibinfo{author}{R.~Kennard},
\newblock \bibinfo{title}{Ridge regression: applications to nonorthogonal problems},
\newblock \bibinfo{journal}{Technometrics} \bibinfo{volume}{12} (\bibinfo{year}{1970}) \bibinfo{pages}{69--82}.
%Type = Article
\bibitem[{Wehbe et~al.(2015)Wehbe, Ramdas, Steorts, and Shalizi}]{Leila_work}
\bibinfo{author}{L.~Wehbe}, \bibinfo{author}{A.~Ramdas}, \bibinfo{author}{R.~Steorts}, \bibinfo{author}{C.~Shalizi},
\newblock \bibinfo{title}{Regularized brain reading with shrinkage and smoothing.},
\newblock \bibinfo{journal}{The annals of applied statistics}  (\bibinfo{year}{2015}) \bibinfo{pages}{1997}.
%Type = Article
\bibitem[{Pasquiou et~al.(2022)Pasquiou, Lakretz, Hale, Thirion, and Pallier}]{NLP_models}
\bibinfo{author}{A.~Pasquiou}, \bibinfo{author}{Y.~Lakretz}, \bibinfo{author}{J.~Hale}, \bibinfo{author}{B.~Thirion}, \bibinfo{author}{C.~Pallier},
\newblock \bibinfo{title}{Neural language models are not born equal to fit brain data, but training helps},
\newblock \bibinfo{journal}{arXiv preprint arXiv:2207.03380}  (\bibinfo{year}{2022}).
%Type = Article
\bibitem[{Conwell et~al.(2022)Conwell, Prince, G., and Konkle}]{newTvision}
\bibinfo{author}{C.~Conwell}, \bibinfo{author}{J.~Prince}, \bibinfo{author}{A.~G.}, \bibinfo{author}{T.~Konkle},
\newblock \bibinfo{title}{Large-scale benchmarking of diverse artificial vision models in prediction of 7t human neuroimaging data},
\newblock \bibinfo{journal}{bioRxiv}  (\bibinfo{year}{2022}).
%Type = Article
\bibitem[{Goldstein et~al.(2022)Goldstein, Ham, Nastase, Zada, Dabush, Aubrey, Schain, Gazula, Feder, Doyle, and Devore}]{GPT_breain}
\bibinfo{author}{A.~Goldstein}, \bibinfo{author}{E.~Ham}, \bibinfo{author}{S.~Nastase}, \bibinfo{author}{Z.~Zada}, \bibinfo{author}{A.~Dabush}, \bibinfo{author}{B.~Aubrey}, \bibinfo{author}{M.~Schain}, \bibinfo{author}{H.~Gazula}, \bibinfo{author}{A.~Feder}, \bibinfo{author}{W.~Doyle}, \bibinfo{author}{S.~Devore},
\newblock \bibinfo{title}{Correspondence between the layered structure of deep language models and temporal structure of natural language processing in the human brain.},
\newblock \bibinfo{journal}{bioRxiv}  (\bibinfo{year}{2022}).
%Type = Article
\bibitem[{Oota et~al.(2022)Oota, Arora, Rowtula, Gupta, and Bapi}]{Lanq-vision}
\bibinfo{author}{S.~Oota}, \bibinfo{author}{J.~Arora}, \bibinfo{author}{V.~Rowtula}, \bibinfo{author}{M.~Gupta}, \bibinfo{author}{R.~Bapi},
\newblock \bibinfo{title}{Visio-linguistic brain encoding.},
\newblock \bibinfo{journal}{preprint arXiv:2204.08261}  (\bibinfo{year}{2022}).
%Type = Article
\bibitem[{Kumar et~al.(2022)Kumar, Sumers, Goldstein, Hasson, Norman, Griffiths, Hawkins, and Nastase}]{bertmodel}
\bibinfo{author}{S.~Kumar}, \bibinfo{author}{T.~Sumers, T.R.and~Yamakoshi}, \bibinfo{author}{A.~Goldstein}, \bibinfo{author}{U.~Hasson}, \bibinfo{author}{K.~Norman}, \bibinfo{author}{T.~Griffiths}, \bibinfo{author}{R.~Hawkins}, \bibinfo{author}{S.~Nastase},
\newblock \bibinfo{title}{Reconstructing the cascade of language processing in the brain using the internal computations of a transformer-based language model.},
\newblock \bibinfo{journal}{bioRxiv}  (\bibinfo{year}{2022}).
%Type = Article
\bibitem[{Lescroart and Gallant(2019)}]{ref_ridge1}
\bibinfo{author}{M.~Lescroart}, \bibinfo{author}{J.~Gallant},
\newblock \bibinfo{title}{Human scene-selective areas represent 3d configurations of surfaces},
\newblock \bibinfo{journal}{Neuron} \bibinfo{volume}{101} (\bibinfo{year}{2019}) \bibinfo{pages}{178--192}.
%Type = Article
\bibitem[{Kell et~al.(2018)Kell, Yamins, Shook, Norman-Haignere, and McDermott}]{ref_ridge2}
\bibinfo{author}{A.~Kell}, \bibinfo{author}{D.~Yamins}, \bibinfo{author}{E.~Shook}, \bibinfo{author}{S.~Norman-Haignere}, \bibinfo{author}{J.~McDermott},
\newblock \bibinfo{title}{A task-optimized neural network replicates human auditory behavior, predicts brain responses, and reveals a cortical processing hierarchy.},
\newblock \bibinfo{journal}{Neuron} \bibinfo{volume}{98} (\bibinfo{year}{2018}) \bibinfo{pages}{630--644}.
%Type = Article
\bibitem[{Jain and Huth(2018)}]{ref_ridge3}
\bibinfo{author}{S.~Jain}, \bibinfo{author}{A.~Huth},
\newblock \bibinfo{title}{Incorporating context into language encoding models for fmri},
\newblock \bibinfo{journal}{Advances in neural information processing systems}  (\bibinfo{year}{2018}) \bibinfo{pages}{31}.
%Type = Article
\bibitem[{Wen et~al.(2018)Wen, Shi, Zhang, Lu, Cao, and Liu}]{VR3}
\bibinfo{author}{H.~Wen}, \bibinfo{author}{J.~Shi}, \bibinfo{author}{Y.~Zhang}, \bibinfo{author}{K.~Lu}, \bibinfo{author}{J.~Cao}, \bibinfo{author}{Z.~Liu},
\newblock \bibinfo{title}{Neural encoding and decoding with deep learning for dynamic natural vision},
\newblock \bibinfo{journal}{Cerebral Cortex} \bibinfo{volume}{28(12))} (\bibinfo{year}{2018}) \bibinfo{pages}{4136--4160}.
%Type = Article
\bibitem[{la~Tour et~al.(2022)la~Tour, Eickenberg, Nunez-Elizalde, and Gallant}]{banded1}
\bibinfo{author}{T.~D. la~Tour}, \bibinfo{author}{M.~Eickenberg}, \bibinfo{author}{A.~O. Nunez-Elizalde}, \bibinfo{author}{J.~L. Gallant},
\newblock \bibinfo{title}{Feature-space selection with banded ridge regression},
\newblock \bibinfo{journal}{NeuroImage} \bibinfo{volume}{264} (\bibinfo{year}{2022}) \bibinfo{pages}{119728}.
%Type = Article
\bibitem[{Seeliger et~al.(2021)Seeliger, Ambrogioni, Güçlütürk, van~den Bulk, Güçlü, and van Gerven}]{kati}
\bibinfo{author}{K.~Seeliger}, \bibinfo{author}{L.~Ambrogioni}, \bibinfo{author}{Y.~Güçlütürk}, \bibinfo{author}{L.~van~den Bulk}, \bibinfo{author}{U.~Güçlü}, \bibinfo{author}{M.~van Gerven},
\newblock \bibinfo{title}{End-to-end neural system identification with neural information flow.},
\newblock \bibinfo{journal}{PLOS Computational Biology} \bibinfo{volume}{17(2)} (\bibinfo{year}{2021}).
%Type = Article
\bibitem[{Kay et~al.(2008)Kay, Naselaris, Prenger, and Gallant}]{vision2}
\bibinfo{author}{K.~Kay}, \bibinfo{author}{T.~Naselaris}, \bibinfo{author}{R.~Prenger}, \bibinfo{author}{J.~Gallant},
\newblock \bibinfo{title}{Identifying natural images from human brain activity},
\newblock \bibinfo{journal}{Nature} \bibinfo{volume}{452} (\bibinfo{year}{2008}) \bibinfo{pages}{352--355}.
%Type = Article
\bibitem[{Kay et~al.(2011)Kay, Naselaris, Prenger, and Gallant}]{vision3}
\bibinfo{author}{K.~Kay}, \bibinfo{author}{T.~Naselaris}, \bibinfo{author}{R.~Prenger}, \bibinfo{author}{J.~Gallant},
\newblock \bibinfo{title}{Reconstructing visual experiences from brain activity evoked by natural movies.},
\newblock \bibinfo{journal}{Current Biology} \bibinfo{volume}{21} (\bibinfo{year}{2011}) \bibinfo{pages}{1641–1646}.
%Type = Article
\bibitem[{Horikawa et~al.(2019)Horikawa, Aoki, Tsukamoto, and Kamitani}]{H_DS}
\bibinfo{author}{T.~Horikawa}, \bibinfo{author}{S.~Aoki}, \bibinfo{author}{M.~Tsukamoto}, \bibinfo{author}{Y.~Kamitani},
\newblock \bibinfo{title}{Characterization of deep neural network features by decodability from human brain activity.},
\newblock \bibinfo{journal}{Scientific data}  (\bibinfo{year}{2019}) \bibinfo{pages}{190012}.
%Type = Article
\bibitem[{Beliy et~al.(2019)Beliy, Gaziv, Hoogi, Strappini, Golan, and Irani}]{BTB}
\bibinfo{author}{R.~Beliy}, \bibinfo{author}{G.~Gaziv}, \bibinfo{author}{A.~Hoogi}, \bibinfo{author}{F.~Strappini}, \bibinfo{author}{T.~Golan}, \bibinfo{author}{M.~Irani},
\newblock \bibinfo{title}{From voxels to pixels and back: Self-supervision in natural-image reconstruction from fmri.},
\newblock \bibinfo{journal}{In Advances in Neural Information Processing Systems}  (\bibinfo{year}{2019}) \bibinfo{pages}{6517--6527}.
%Type = Article
\bibitem[{Qiao et~al.(2019)Qiao, Chen, Wang, Zhang, Zeng, Tong, and Yan}]{LSTM1}
\bibinfo{author}{K.~Qiao}, \bibinfo{author}{J.~Chen}, \bibinfo{author}{L.~Wang}, \bibinfo{author}{C.~Zhang}, \bibinfo{author}{L.~Zeng}, \bibinfo{author}{L.~Tong}, \bibinfo{author}{B.~Yan},
\newblock \bibinfo{title}{Category decoding of visual stimuli from human brain activity using a bidirectional recurrent neural network to simulate bidirectional information flows in human visual cortices},
\newblock \bibinfo{journal}{Frontiers in neuroscience}  (\bibinfo{year}{2019}).
%Type = Article
\bibitem[{Shen et~al.(2019)Shen, Dwivedi, Majima, Horikawa, and Kamitani}]{IM_1}
\bibinfo{author}{G.~Shen}, \bibinfo{author}{K.~Dwivedi}, \bibinfo{author}{K.~Majima}, \bibinfo{author}{T.~Horikawa}, \bibinfo{author}{Y.~Kamitani},
\newblock \bibinfo{title}{End-to-end deep image reconstruction from human brain activity.},
\newblock \bibinfo{journal}{Frontiers in computational neuroscience} \bibinfo{volume}{13} (\bibinfo{year}{2019}) \bibinfo{pages}{21}.
%Type = Article
\bibitem[{Naselaris et~al.(2021)Naselaris, Allen, and Kay}]{Dataset_p}
\bibinfo{author}{T.~Naselaris}, \bibinfo{author}{E.~Allen}, \bibinfo{author}{K.~Kay},
\newblock \bibinfo{title}{Extensive sampling for complete models of individual brains. current opinion in behavioral sciences},
\newblock \bibinfo{journal}{bioRxiv} \bibinfo{volume}{40} (\bibinfo{year}{2021}) \bibinfo{pages}{45--51}.
%Type = Article
\bibitem[{Chang et~al.(2019)Chang, Pyles, Marcus, Gupta, Tarr, and Aminoff}]{Bold5000}
\bibinfo{author}{N.~Chang}, \bibinfo{author}{J.~Pyles}, \bibinfo{author}{A.~Marcus}, \bibinfo{author}{A.~Gupta}, \bibinfo{author}{M.~Tarr}, \bibinfo{author}{E.~Aminoff},
\newblock \bibinfo{title}{Bold5000: a public fmri dataset while viewing 5000 visual images.},
\newblock \bibinfo{journal}{Scientific data}  (\bibinfo{year}{2019}) \bibinfo{pages}{1--18}.
%Type = Article
\bibitem[{Allen et~al.(2021)Allen, St-Yves, Wu, Breedlove, Dowdle, Caron, Pestilli, Charest, Hutchinson, Naselaris, and Kay}]{NSD_dataset}
\bibinfo{author}{E.~Allen}, \bibinfo{author}{G.~St-Yves}, \bibinfo{author}{Y.~Wu}, \bibinfo{author}{J.~Breedlove}, \bibinfo{author}{L.~Dowdle}, \bibinfo{author}{B.~Caron}, \bibinfo{author}{F.~Pestilli}, \bibinfo{author}{I.~Charest}, \bibinfo{author}{J.~Hutchinson}, \bibinfo{author}{T.~Naselaris}, \bibinfo{author}{K.~Kay},
\newblock \bibinfo{title}{A massive 7t fmri dataset to bridge cognitive and computational neuroscience.},
\newblock \bibinfo{journal}{bioRxiv}  (\bibinfo{year}{2021}).
%Type = Article
\bibitem[{ratton et~al.(2018)ratton, Laumann, Nielsen, Greene, Gordon, Gilmore, Nelson, Coalson, Snyder, Schlaggar, and Dosenbach}]{Gratton}
\bibinfo{author}{C.~ratton}, \bibinfo{author}{T.~Laumann}, \bibinfo{author}{A.~Nielsen}, \bibinfo{author}{D.~Greene}, \bibinfo{author}{E.~Gordon}, \bibinfo{author}{A.~Gilmore}, \bibinfo{author}{S.~Nelson}, \bibinfo{author}{R.~Coalson}, \bibinfo{author}{A.~Snyder}, \bibinfo{author}{B.~Schlaggar}, \bibinfo{author}{N.~Dosenbach},
\newblock \bibinfo{title}{Functional brain networks are dominated by stable group and individual factors, not cognitive or daily variation.},
\newblock \bibinfo{journal}{Neuron}  (\bibinfo{year}{2018}) \bibinfo{pages}{439--452}.
%Type = Misc
\bibitem[{Cneuromod(2021)}]{cneuromod}
\bibinfo{author}{Cneuromod}, \bibinfo{title}{Cneuromod dataset}, \bibinfo{howpublished}{\url{https://www.cneuromod.ca/gallery/datasets/}}, \bibinfo{year}{2021}.
%Type = Article
\bibitem[{Pedregosa et~al.(2011)Pedregosa, Varoquaux, Gramfort, Michel, Thirion, Grisel, Blondel, Prettenhofer, Weiss, Dubourg et~al.}]{pedregosa2011scikit}
\bibinfo{author}{F.~Pedregosa}, \bibinfo{author}{G.~Varoquaux}, \bibinfo{author}{A.~Gramfort}, \bibinfo{author}{V.~Michel}, \bibinfo{author}{B.~Thirion}, \bibinfo{author}{O.~Grisel}, \bibinfo{author}{M.~Blondel}, \bibinfo{author}{P.~Prettenhofer}, \bibinfo{author}{R.~Weiss}, \bibinfo{author}{V.~Dubourg}, et~al.,
\newblock \bibinfo{title}{Scikit-learn: Machine learning in python},
\newblock \bibinfo{journal}{the Journal of machine Learning research} \bibinfo{volume}{12} (\bibinfo{year}{2011}) \bibinfo{pages}{2825--2830}.
%Type = Article
\bibitem[{Xianyi et~al.(2012)Xianyi, Qian, and Yunquan}]{OpenBLAS}
\bibinfo{author}{Z.~Xianyi}, \bibinfo{author}{W.~Qian}, \bibinfo{author}{Z.~Yunquan},
\newblock \bibinfo{title}{Model-driven level 3 blas performance optimization on loongson 3a processor.},
\newblock \bibinfo{journal}{IEEE 18th international conference on parallel and distributed systems}  (\bibinfo{year}{2012}) \bibinfo{pages}{1--18}.
%Type = Article
\bibitem[{Wang et~al.(2014)Wang, Zhang, Shen, Zhang, Lu, Wu, and Wang}]{MKL}
\bibinfo{author}{E.~Wang}, \bibinfo{author}{Q.~Zhang}, \bibinfo{author}{B.~Shen}, \bibinfo{author}{G.~Zhang}, \bibinfo{author}{X.~Lu}, \bibinfo{author}{Q.~Wu}, \bibinfo{author}{Y.~Wang},
\newblock \bibinfo{title}{Intel math kernel library. in high-performance},
\newblock \bibinfo{journal}{Computing on the Intel® Xeon Phi™}  (\bibinfo{year}{2014}) \bibinfo{pages}{167--188}.
%Type = Article
\bibitem[{Rocklin(2015)}]{Dask}
\bibinfo{author}{M.~Rocklin},
\newblock \bibinfo{title}{Dask: Parallel computation with blocked algorithms and task scheduling},
\newblock \bibinfo{journal}{In Proceedings of the 14th python in science conferenc} \bibinfo{volume}{130} (\bibinfo{year}{2015}) \bibinfo{pages}{136}.
%Type = Article
\bibitem[{Boyle and Pinsard(2021)}]{cneuromod1}
\bibinfo{author}{J.~Boyle}, \bibinfo{author}{e.~a. Pinsard, B.},
\newblock \bibinfo{title}{The courtois project on neuronal modeling - 2021 data release},
\newblock \bibinfo{journal}{Poster 2224 was presented at the 2021 Annual Meeting of the Organization for Human Brain Mapping held virtually}  (\bibinfo{year}{2021}).
%Type = Article
\bibitem[{Setsompop et~al.(2012)Setsompop, Cohen-Adad, Gagoski, Raij, Yendiki, Keil, Wedeen, and Wald}]{Setsompop}
\bibinfo{author}{K.~Setsompop}, \bibinfo{author}{J.~Cohen-Adad}, \bibinfo{author}{B.~Gagoski}, \bibinfo{author}{T.~Raij}, \bibinfo{author}{A.~Yendiki}, \bibinfo{author}{B.~Keil}, \bibinfo{author}{V.~Wedeen}, \bibinfo{author}{L.~Wald},
\newblock \bibinfo{title}{Improving diffusion mri using simultaneous multi-slice echo planar imaging.},
\newblock \bibinfo{journal}{Neuroimage} \bibinfo{volume}{63} (\bibinfo{year}{2012}) \bibinfo{pages}{569--580}.
%Type = Article
\bibitem[{Xu et~al.(2013)Xu, Moeller, Auerbach, Strupp, Smith, Feinberg, Yacoub, and Uğurbil}]{Xu}
\bibinfo{author}{J.~Xu}, \bibinfo{author}{S.~Moeller}, \bibinfo{author}{E.~Auerbach}, \bibinfo{author}{J.~Strupp}, \bibinfo{author}{S.~Smith}, \bibinfo{author}{D.~Feinberg}, \bibinfo{author}{E.~Yacoub}, \bibinfo{author}{K.~Uğurbil},
\newblock \bibinfo{title}{Improving diffusion mri using simultaneous multi-slice echo planar imaging.},
\newblock \bibinfo{journal}{NeuroimageT} \bibinfo{volume}{83} (\bibinfo{year}{2013}) \bibinfo{pages}{991--1001}.
%Type = Article
\bibitem[{Van~Essen and Glasser(2016)}]{Glasser}
\bibinfo{author}{D.~Van~Essen}, \bibinfo{author}{M.~Glasser},
\newblock \bibinfo{title}{The human connectome project: Progress and prospects. in cerebrum: the dana forum on brain science},
\newblock \bibinfo{journal}{Dana Foundation} \bibinfo{volume}{63} (\bibinfo{year}{2016}).
%Type = Article
\bibitem[{Esteban et~al.(2019)Esteban, Markiewicz, Blair, Moodie, Isik, Erramuzpe, Kent, Goncalves, DuPre, Snyder, and Oya}]{Esteban}
\bibinfo{author}{O.~Esteban}, \bibinfo{author}{C.~Markiewicz}, \bibinfo{author}{R.~Blair}, \bibinfo{author}{C.~Moodie}, \bibinfo{author}{A.~Isik}, \bibinfo{author}{A.~Erramuzpe}, \bibinfo{author}{J.~Kent}, \bibinfo{author}{M.~Goncalves}, \bibinfo{author}{E.~DuPre}, \bibinfo{author}{M.~Snyder}, \bibinfo{author}{H.~Oya},
\newblock \bibinfo{title}{fmriprep: a robust preprocessing pipeline for functional mri. nature methods},
\newblock \bibinfo{journal}{Neuroimage} \bibinfo{volume}{16} (\bibinfo{year}{2019}) \bibinfo{pages}{111--116}.
%Type = Article
\bibitem[{Abraham et~al.(2014)Abraham, Pedregosa, Eickenberg, Gervais, Mueller, Kossaifi, Gramfort, Thirion, and Varoquaux}]{ML_neuroimaging}
\bibinfo{author}{A.~Abraham}, \bibinfo{author}{F.~Pedregosa}, \bibinfo{author}{M.~Eickenberg}, \bibinfo{author}{P.~Gervais}, \bibinfo{author}{A.~Mueller}, \bibinfo{author}{J.~Kossaifi}, \bibinfo{author}{A.~Gramfort}, \bibinfo{author}{B.~Thirion}, \bibinfo{author}{G.~Varoquaux},
\newblock \bibinfo{title}{Machine learning for neuroimaging with scikit-learn},
\newblock \bibinfo{journal}{Frontiers in neuroinformatics}  (\bibinfo{year}{2014}) \bibinfo{pages}{14}.
%Type = Article
\bibitem[{Urchs et~al.(2019)Urchs, Armoza, Moreau, Benhajali, St-Aubin, Orban, and Bellec}]{MIST}
\bibinfo{author}{S.~Urchs}, \bibinfo{author}{J.~Armoza}, \bibinfo{author}{C.~Moreau}, \bibinfo{author}{Y.~Benhajali}, \bibinfo{author}{J.~St-Aubin}, \bibinfo{author}{P.~Orban}, \bibinfo{author}{P.~Bellec},
\newblock \bibinfo{title}{Mist: A multi-resolution parcellation of functional brain networks},
\newblock \bibinfo{journal}{MNI Open Research} \bibinfo{volume}{1} (\bibinfo{year}{2019}) \bibinfo{pages}{3}.
%Type = Article
\bibitem[{Conwell et~al.(2022)Conwell, Prince, Kay, Alvarez, and Konkle}]{conwell2022pressures}
\bibinfo{author}{C.~Conwell}, \bibinfo{author}{J.~S. Prince}, \bibinfo{author}{K.~N. Kay}, \bibinfo{author}{G.~A. Alvarez}, \bibinfo{author}{T.~Konkle},
\newblock \bibinfo{title}{What can 1.8 billion regressions tell us about the pressures shaping high-level visual representation in brains and machines?},
\newblock \bibinfo{journal}{BioRxiv}  (\bibinfo{year}{2022}).
%Type = Article
\bibitem[{Simonyan and Zisserman(2014)}]{VGG16_ref}
\bibinfo{author}{K.~Simonyan}, \bibinfo{author}{A.~Zisserman},
\newblock \bibinfo{title}{Very deep convolutional networks for large-scale image recognition},
\newblock \bibinfo{journal}{arXiv preprint arXiv}  (\bibinfo{year}{2014}).
%Type = Article
\bibitem[{Deng et~al.(2009)Deng, Dong, Socher, and Fei-Fei}]{ImageNet}
\bibinfo{author}{J.~Deng}, \bibinfo{author}{W.~Dong}, \bibinfo{author}{R.~L. Socher}, \bibinfo{author}{L.~Fei-Fei},
\newblock \bibinfo{title}{Imagenet: A large-scale hierarchical image database.},
\newblock \bibinfo{journal}{IEEE conference on computer vision and pattern recognition}  (\bibinfo{year}{2009}) \bibinfo{pages}{248--255}.
%Type = Article
\bibitem[{Simonyan and Zisserman(2014)}]{vgg16}
\bibinfo{author}{K.~Simonyan}, \bibinfo{author}{A.~Zisserman},
\newblock \bibinfo{title}{Very deep convolutional networks for large-scale image recognition},
\newblock \bibinfo{journal}{arXiv preprint arXiv}  (\bibinfo{year}{2014}).
%Type = Article
\bibitem[{Logothetis et~al.(2001)Logothetis, Pauls, Augath, Trinath, and Oeltermann}]{Neurophysiological}
\bibinfo{author}{N.~Logothetis}, \bibinfo{author}{J.~Pauls}, \bibinfo{author}{M.~Augath}, \bibinfo{author}{T.~Trinath}, \bibinfo{author}{A.~Oeltermann},
\newblock \bibinfo{title}{Neurophysiological investigation of the basis of the fmri signal},
\newblock \bibinfo{journal}{nature}  (\bibinfo{year}{2001}) \bibinfo{pages}{150--157}.
%Type = Inproceedings
\bibitem[{Moritz et~al.(2018)Moritz, Nishihara, Wang, Tumanov, Liaw, Liang, Elibol, Yang, Paul, Jordan et~al.}]{moritz2018ray}
\bibinfo{author}{P.~Moritz}, \bibinfo{author}{R.~Nishihara}, \bibinfo{author}{S.~Wang}, \bibinfo{author}{A.~Tumanov}, \bibinfo{author}{R.~Liaw}, \bibinfo{author}{E.~Liang}, \bibinfo{author}{M.~Elibol}, \bibinfo{author}{Z.~Yang}, \bibinfo{author}{W.~Paul}, \bibinfo{author}{M.~I. Jordan}, et~al.,
\newblock \bibinfo{title}{Ray: A distributed framework for emerging $\{$AI$\}$ applications},
\newblock in: \bibinfo{booktitle}{13th USENIX symposium on operating systems design and implementation (OSDI 18)}, pp. \bibinfo{pages}{561--577}.

\end{thebibliography}

\subsection{Appendix}
\label{Appendix1}
\begin{table}[ht]
%\hspace*{-20 cm}
\centering
\caption{\small Key Parameters of VGG16 (Keras Model)}
\small

\begin{tabular}{cccccc}
  Layer & Num. of & Activation & Size of  & Parameters \\
  & Kernels & Size & Kernels & (M) \\
  \hline
  Input & - & 224x224x3 & - & - \\
  \hline
  block1\_conv1 & 64 & 224x224x64 & 3x3 & 1792 \\
  block1\_conv2 & 64 & 224x224x64 & 3x3 & 36928 \\
  block1\_pool & - & 112x112x64 & 2x2 & - \\
  \hline
  block2\_conv1 & 128 & 112x112x128 & 3x3 & 73856 \\
  block2\_conv2 & 128 & 112x112x128 & 3x3 & 147584 \\
  block2\_pool & - & 56x56x128 & 2x2 & - \\
  \hline
  block3\_conv1 & 256 & 56x56x256 & 3x3 & 295168 \\
  block3\_conv2 & 256 & 56x56x256 & 3x3 & 590080 \\
  block3\_conv3 & 256 & 56x56x256 & 3x3 & 590080 \\
  block3\_pool & - & 28x28x256 & 2x2 & - \\
  \hline
  block4\_conv1 & 512 & 28x28x512 & 3x3 & 1180160 \\
  block4\_conv2 & 512 & 28x28x512 & 3x3 & 2359808 \\
  block4\_conv3 & 512 & 28x28x512 & 3x3 & 2359808 \\
  block4\_pool & - & 14x14x512 & 2x2 & - \\
  \hline
  block5\_conv1 & 512 & 14x14x512 & 3x3 & 2359808 \\
  block5\_conv2 & 512 & 14x14x512 & 3x3 & 2359808 \\
  block5\_conv3 & 512 & 14x14x512 & 3x3 & 2359808 \\
  block5\_pool & - & 7x7x512 & 2x2 & - \\
  \hline
  Flatten & - & 25088 & - & - \\
  FC1 & - & 4096 & - & 102764544 \\
  FC2 & - & 4096 & - & 16781312 \\
  predictions & - & 1000 (output) & - & 4097000 \\
  \hline
  Total & - & - & - & 138,357,544 \\
\end{tabular}
\label{tab:vgg16_keras}
\end{table}

\end{document}